\documentclass[5p,final]{elsarticle}
\usepackage{graphicx}
\usepackage{siunitx}
\usepackage[ruled,linesnumbered]{algorithm2e}
\SetKwInput{KwInput}{Input}                
\SetKwInput{KwOutput}{Output}              
\usepackage{physics}
\usepackage{color}
\usepackage{amsmath}
\usepackage{multirow}
\newcommand{\etal}{\textit{et al.}}
\journal{Microelectronics Journal}

\begin{document}

\begin{frontmatter}

\title{Performance Evaluation of Evolutionary Algorithms for Analog Integrated Circuit Design Optimisation}

\author[inst1]{Ria Rashid\corref{cor1}}
\cortext[cor1]{Corresponding Author}
\ead{ria183422005@iitgoa.ac.in}
\author[inst1]{Gopavaram Raghunath}
\author[inst1]{Vasant Badugu}
\author[inst1]{Nandakumar Nambath}

\affiliation[inst1]{organization={School of Electrical Sciences, Indian Institute of Technology Goa},
            city={Ponda},
            postcode={403401}, 
            state={Goa},
            country={India}}

\begin{abstract}

An automated sizing approach for analog circuits using evolutionary algorithms is presented in this paper. A targeted search of the search space has been implemented using a particle generation function and a repair-bounds function that has resulted in faster convergence to the optimal solution. The algorithms are tuned and modified to converge to a better optimal solution with less standard deviation for multiple runs compared to standard versions. Modified versions of the artificial bee colony optimisation algorithm, genetic algorithm, grey wolf optimisation algorithm, and particle swarm optimisation algorithm are tested and compared for the optimal sizing of two operational amplifier topologies. An extensive performance evaluation of all the modified algorithms showed that the modifications have resulted in consistent performance with improved convergence for all the algorithms. The implementation of parallel computation in the algorithms has reduced run time. Among the considered algorithms, the modified artificial bee colony optimisation algorithm gave the most optimal solution with consistent results across multiple runs.
\end{abstract}


\end{frontmatter}


\section{Introduction}

 Automation in analog integrated circuit (IC) design has yet to advance to the same extent as digital IC design, mainly due to highly complex analog design procedure\cite{1,2}. The optimal design of analog circuits depends on several design specifications, including gain, phase margin, power, area, and bandwidth \cite{3}. Analog circuits are more often designed manually, where the experience of analog designers plays a significant role. Analog circuit automation continues to be an active area of research due to the complexities and time involved in manually designing analog circuits \cite{4,5,6}.
 
Different approaches have been reported in the literature to automate the device sizing aspect of the analog IC design\cite{7,8,9}. One such approach is the simulation-based optimisation method\cite{10,11,12}. In this approach, an optimisation algorithm is used to arrive at an analog IC's optimal design parameters satisfying the target specifications. Electronic design automation tools are used in combination with optimisation algorithms to evaluate the solutions generated by the algorithm. Several optimisation algorithms, including simulated annealing\cite{13,14}, multiple start points algorithm\cite{15}, Bayesian optimisation\cite{16,17}, artificial intelligence based optimisation \cite{18}, machine learning-assisted optimisation\cite{19,20}, artificial neural network based optimisation\cite{21}, surrogate based optimisation \cite{22,23,24}, gravitational search algorithm \cite{25} and evolutionary algorithms (EAs)\cite{26,27,28,29,30} have been reported in the literature to optimise the analog circuit sizing problem.
 
Several studies have used evolutionary computation techniques for optimising analog circuits. A performance evaluation of different EAs, namely, genetic algorithm (GA), particle swarm optimisation (PSO), and artificial bee colony optimisation (ABCO), was carried out for the optimal design of analog filters in \cite{31}. In \cite{32}, PSO was reported to be used for optimising a low-noise amplifier and a second-generation current conveyor (CC). The results showed the effectiveness of PSO in optimising analog circuits. Hierarchical PSO, a modified version of the standard PSO, was reported in \cite{33} for the automatic sizing of low-power analog circuits and was found to generate circuits with better performance than manually designed circuits. PSO was utilized for the optimal sizing of analog circuits with high accuracy and reduced computational time in \cite{34}. In \cite{35}, ABCO was reported to have better performance than other reported studies for the optimisation of a folded cascode operational transconductance amplifier (OTA). The performance of the ABCO for an inverter design considering propagation delays was studied in \cite{36}. A tool that can synthesize complementary metal oxide semiconductor (CMOS) operational amplifiers (op-amps) called DARWIN, based on GA, was reported in \cite{37}. An optimisation system based on a non-dominated sorting GA and multi-objective EA based on decomposition was reported in \cite{38} and was applied for the optimisation of two CC circuits. In \cite{39}, a method for device sizing of an OTA using GA was presented, with the simulation results confirming the efficiency of GA in finding the optimal design. An adaptive immune GA was reported in \cite{40} for the optimal design of a low-pass filter. A rule-guided GA for analog circuit optimisation reported in \cite{41} was found to have faster convergence when compared to standard GA.

The main drawback of the simulation-based approach is that the circuit simulations can get computationally expensive; hence, there is a need to reduce the number of evaluations to find the optimal design and speed up the convergence of algorithms. In this work, we have proposed a simulation-based analog circuit design optimisation methodology using different evolutionary algorithms for faster convergence and run time reduction. In the proposed method, characteristics of the analog circuit under study are considered for setting the search space boundaries during the execution of the optimisation algorithms. This is implemented using a particle generation function, which generates candidate solutions for the initial population, and the repair-bounds function, which ensures each candidate solution is in a feasible region in the search space after every update. This helps the algorithms' targeted search in the search space, resulting in fewer spice calls and reduced execution time. Parallel computation has been implemented in all the optimisation algorithms for lesser run time. Since the standard versions of the considered evolutionary algorithms converged to solutions with high standard deviation for multiple runs, all the algorithms have been modified and tuned for convergence to better optimal solutions with lesser standard deviation for multiple runs. This study aims to find the best evolutionary algorithm for analog design from among the algorithms considered in this study by conducting a performance analysis through exhaustive simulations. In our prior studies, a modified version of PSO was applied for the area optimisation of a differential amplifier\cite{42} and a two-stage op-amp\cite{43}. 

The remainder of this paper is organized as follows. Section 2 describes the proposed automated sizing approach. Different algorithms used for the optimisation are described in Section 3. The formulation of the optimisation problems is summarised in section 4. Section 5 reports the simulation results and discussion, and the conclusion is provided in Section 6.
 
\section{Automated Sizing Approach Using Evolutionary Algorithms}

 If analog circuit design is considered as an optimisation problem with various circuit specifications as constraints, design parameters as decision variables, and the circuit parameter to be optimised as the objective function, optimisation algorithms can be used to obtain optimal design parameters. Objective functions and the decision variables are chosen depending on the circuit under consideration. Bounds for the decision variables form the search space for the analog circuit optimisation problem. According to the optimisation algorithm used, different sets of decision variables are moved in the search space in search of the optimal solution. The primary aim of our work is to optimise analog circuits in terms of transistor sizing. Therefore, the circuit area is chosen as the objective function to be optimised in the analog circuits considered in this study. 
 
 The proposed optimisation methodology is divided into three phases: population generator function, survivability test, and optimisation algorithm with repair bounds.
 
\subsection{Population Generator Function}

 Any population-based metaheuristic optimisation algorithm requires an initial population, where each member is a potential solution for the optimisation problem. In the optimisation methodology proposed here, a population generator function (PGF) generates the members for the initial population. This function is framed based on the mathematical modeling of analog circuits.
 
 In analog circuits, operation regions of all the transistors are to be ensured. Also, the optimised design must meet all the constraints of the targeted circuit specifications. By mathematically modeling these conditions of circuit specifications and the correct region of operation, it is possible to find the feasible bounds of values for each circuit design parameter where these conditions are met. Each design parameter also has the bounds that form the search space for the optimisation problem. An intersection of these two bounds is considered for each design parameter. A random value is picked from the respective bounds for each circuit design parameter to form a potential solution in the initial population.
 
 In PGF, for the mathematical modeling of the analog circuit under study, lower-order, simple models of the transistors are used. Hence, the result obtained from the modeling of transistors lacks accuracy compared to any circuit simulator, which employs more accurate and higher-order transistor models. Even then, a solution generated using PGF has a higher probability of satisfying all the conditions than a solution generated randomly from within the specified bounds. PGF can substantially reduce the time required to generate a member satisfying all the constraints for the initial population.
 
\subsection{Survivability Test}

 The survivability test is used to test whether all the candidate solutions in the population satisfy the optimisation problem's constraints. As explained in the above section, the PGF uses less accurate models for generating a candidate solution; hence, it cannot be ensured that the generated solution satisfies all the constraints. Therefore, for all the solutions generated by the PGF, a survivability test is carried out using circuit simulator software to get more accurate results. The survivability test is also carried out for every new solution generated during the execution of the optimisation algorithm. This additional check ensures that all the updated solutions in the population in every iteration satisfy the constraints. Ngspice is used for the circuit simulations in this test. Different analyses, namely, DC operating point analysis, AC analysis, noise analysis, etc., are performed as part of the survivability test.
 
\subsection{Optimisation Algorithm with Repair-Bounds}

 Four different EAs, namely, ABCO, GA, gray wolf optimisation (GWO), and PSO, have been considered in this work due to their broad applicability in various engineering fields. Modified versions of the algorithms are used to optimise the area of two op-amp topologies: a two-stage op-amp and a folded cascode op-amp. Standard versions of the algorithms mentioned above were used initially for optimisation. Multiple runs were carried out for the same circuit design to test the robustness of the standard algorithms. Although the standard algorithms converged to reasonable solutions, the standard deviations in the objective function values across multiple runs were significant. All the algorithms were then modified to obtain better optimal solutions with lesser standard deviation, with a substantial reduction in the number of spice calls and simulation times. The modified algorithms are explained in detail in section 3. Since all the candidate solutions in a population are independent of each other in the algorithms considered in this study, codes have been parallelised for faster execution.
 
 Whenever the optimisation algorithm updates a candidate solution, it is passed onto a repair-bounds function before invoking the survivability test. The repair-bounds function calculates the feasible bounds for each decision variable in the same manner as the PGF. The primary objective of the repair-bounds function is to check whether any decision variables have gone beyond the calculated bounds. In case the value goes beyond the bounds, that decision variable is set to the lower or the upper bound, whichever is the nearest. This has resulted in significantly reducing the number of spice calls per candidate solution during the execution of the algorithms.
 
\section{Evolutionary Algorithms}

\subsection{Modified Artificial Bee Colony Optimisation (MABCO)}

{\SetAlgoNoLine%
\begin{algorithm}[]
\DontPrintSemicolon
  \KwInput{$N$, $D$, $max_{ite}$, $max_{count}$}
  \KwOutput{Best food source (Optimal solution)}
  \BlankLine
  Generate initial $N$ food sources, $\vb*X$, using PGF.\;
  Calculate fitness value, $f(\vb*X)$\ and set $trial[N]=0$.\;
  \For{$j\gets1$ \KwTo $max_{ite}$}{
    Update $limit_{j}$ and $dim_{j}$ using equations, (\ref{eq3}) and (\ref{eq4}). Set $count=0$.\;
    \For{$i\gets1$ \KwTo $N$}{
    \While{true}{
    Generate new food source, $\vb*v_{i,j}$, using equation (\ref{eq2}) in employed bee phase.\;
    Perform greedy selection.\;
    Conduct survivability test for food source.\;
    \eIf{survivability test passed}{Set $count=0$ and break.\;}{$count=count+1$.}
    \If{$count=max_{count}$}{Retain the old food source, $\vb*x_{i,j}$.\;
    Set $count=0$ and break.\;}
    }
    \If{unable to find better food source}{Update $trial[i]=trial[i]+1$.}
    \While{true}{
    Select a food source, $\vb*x_{i,j(o)}$ using a roulette wheel selection scheme.\;
    Generate new food source, $\vb*v_{i,j(o)}$, using equation (\ref{eq2}) in onlooker bee phase.\;
    Perform greedy selection.\;
    Conduct survivability test for food source.\;
    \eIf{survivability test passed}{Set $count=0$ and break.\;}{$count=count+1$.}
    \If{$count=max_{count}$}{Retain the old food source, $\vb*x_{i,j(o)}$.\;
    Set $count=0$ and break.\;}
    }
    \If{$trial[i]\geq limit_{ite}$}{
    Find new food source from $\vb*x_{i,j}$ and the best food source.\;
    Find another new food source from two randomly picked food sources.\;
    Replace $\vb*x_{i,j}$ with the best out of the above two solutions in the scout bee phase.\;
    Set $trial[i] = 0$.\;}
    }
    Memorize the best food source.
    }
\caption{MABCO.}
\label{alg1}
\end{algorithm}}%

 ABCO is a population-based metaheuristic algorithm proposed by Karaboga in 2007\cite{44}. It is based on honey bees' intelligent organizational nature and foraging behavior when seeking a quality food source. The standard version of the ABCO is explained in detail in \cite{45}.

 The following modifications are made to standard ABCO to obtain faster convergence and consistent results for multiple runs. Let the population size be $N$ and the dimension of the search space be $D$. The limit parameter of the ABCO is denoted as $limit$. The PGF generates the initial food sources. The $i^\text{th}$ food source in the $j^\text{th}$ iteration can be represented as
  \begin{equation}
    \label{eq1}\vb*x_{i,j} = [x_{1}, x_{2},x_{3}, \cdots, x_{D}]_{i,j}.
  \end{equation} 
 In the employed bee and the onlooker bee phase, all the elements, $x_1,x_2,x_3,\cdots,x_D$, constituting the food source, $x_{i,j}$, are updated to find a new food source, using the equation,
  \begin{equation}
 \label{eq2}
    \vb*v_{i,j} = \vb*x_{i,j} + rand(-1,1)\times(\vb*x_{i,j}-\vb*x_{k,j})
 \end{equation}
 where $i,k \in (1,2...,N)$ and $i\neq k$, in the standard ABCO algorithm. In the modified version, the number of elements in the food source that are updated is kept to a high value in the initial stages and a lower value in the final stages to facilitate faster convergence of the algorithm. Also, the elements which have to be updated are picked randomly. 
 
 The limit parameter value is kept constant in the standard ABCO in the scout bee phase. It is observed that as the limit value is decreased, the algorithm enters the scout bee phase more often. The lower limit value increases the randomness in the population towards the final stages leading to slow convergence. If the limit value is kept high, the scout bee phase is executed rarely, leading to less randomness in the population resulting in premature convergence. Hence, in the MABCO, the limit is initially set to a higher value to aid global exploration and lower values in the final stages to facilitate faster convergence. The limit parameter, $limit_{ite}$, and the number of dimensions to be updated, $dim_{ite}$, for the $ite^\text{th}$ iteration is found using the equations,
  \begin{equation}
  \begin{split}
    \label{eq3}
    limit_{ite}  = integer[limit_{min}+(1-\frac{ite}{ite_{max}})\times\\ (limit_{max}-limit_{min})] ~\text{and}
    \end{split}
  \end{equation}
  \begin{equation}
    \label{eq4}
    dim_{ite}  = max[D \times (1-\frac{ite}{ite_{max}}),1]
  \end{equation}
 where $D$ is the dimension of the search space, $ite_{max}$ is the maximum number of iterations, $limit_{min}$ is the lower bound, and $limit_{max}$ is the upper bound of the limit parameter. 
 
 In the standard ABCO, once the limit is reached in the scout bee phase, the exhausted food source is eliminated and replaced with a new random food source. This aids in the global exploration of the search space. The algorithm generally enters the scout bee phase more often in the final stages. If we replace the exhausted solution with a completely random solution, there is a high probability that the random solution will be away from the best solution, leading to slower convergence. Furthermore, the experience of the exhausted solution is not being used to generate the new solution. We have modified this stage of the standard ABCO. Rather than eliminating the exhausted solution, a new potential solution is created from the exhausted solution, and the best solution achieved so far by a crossover operation. Another candidate solution is generated by a crossover operation between two randomly selected existing solutions. The crossover operation used here is the same as in GA. The crossover point is picked randomly. This process is repeated till two new candidate solutions are obtained. The better of the two candidate solutions is selected to replace the exhausted solution. These steps are carried out throughout the iterations till the stopping criteria are met. A pseudo-code of the MABCO is given in Algorithm \ref{alg1}.

\subsection{Modified Genetic Algorithm (MGA)}

 GA is a metaheuristic optimisation method based on the theory of natural evolution, introduced by John Holland in 1971\cite{46}. This algorithm is governed by the concept of the survival of the fittest. The standard version of the algorithm is explained in detail in \cite{47}.
 
 {\SetAlgoNoLine%
 \begin{algorithm}[t!]
\DontPrintSemicolon
 \BlankLine
  \KwInput{$N$, $D$, $gen_{max}$}
  \KwOutput{Best individual (Optimal solution)}
  \BlankLine
  Generate initial population, $\vb*X$, using PGF.\;
   \For{$i\gets1$ \KwTo $N$}{
    Calculate fitness value, $f(\vb*x_i)$\;
    }
  \For{$gen\gets1$ \KwTo $gen_{max}$}{
    \For{$i\gets1$ \KwTo $N$}{
    \While{true}{
    Randomly pick two parents and the crossover point.\;
    Generate crossover offspring, $\vb*x_{c,i}$.\;
    \If{offspring passes survivability test}{break.}
    }
    \While{true}{
    Find search space parameter, $\alpha$ and outer bounds using equations (\ref{eq5}) to (\ref{eq7}).\;
    Randomly pick $mutation_{count}$ and $mutation_{no}$.\;
    Generate mutated crossed offspring, $\vb*x_{cm,i}$.\;
    \If{offspring passes survivability test}{break.}
    }
    \While{true}{
    Find search space parameter and outer bounds using equations (\ref{eq5}) to (\ref{eq7}).\;
    Randomly pick a parent, $mutation_{count}$ and $mutation_{no}$.\;
    Generate mutated parent offspring, $\vb*x_{pm,i}$.\;
    \If{offspring passes survivability test}{break.}
    }
    }
    Pool together the parents, $\vb*X$ and offspring, $\vb*X_{c}$, $\vb*X_{cm}$, and $\vb*X_{pm}$, to form $\vb*X_{new}$.\;
    \For{$i\gets1$ \KwTo $4N$}{
    Calculate fitness, $f(\vb*x_{new,i})$.\;
    }
    Select fittest $N$ individuals to form new population, $\vb*X$.\;
    Update best individual.\;
    }
\caption{MGA.}
\label{alg2}
\end{algorithm}}%

  In MGA, the PGF generates the initial population. After the selection based on the fitness scores, parents are randomly chosen to create offspring in each generation using the crossover operation. The crossover point is randomly selected. For each crossover offspring produced, a mutated offspring is also generated. This study uses random mutation for the mutation operation. The number and which genes to be mutated are picked randomly. This random mutation has resulted in faster convergence and avoided premature algorithm convergence. Some parents are also randomly picked to create mutated offspring. A survivability test is carried out for all the offspring generated. If any generated offspring fails the test, the steps are repeated until a suitable offspring is generated. The parents, crossover offspring, crossover mutated offspring, and parent mutated offspring are then pooled together, and the fittest among them are carried over to the next generation.
 
 We have introduced a search space parameter, $\alpha$, in the algorithm. $\alpha$ is linearly decreased from $\alpha_{max}$ to $\alpha_{min}$ over the generations. This parameter is used to limit the search space for the mutation operation. The varying values of $\alpha$ in the initial and later generations aid global and local explorations, respectively. This parameter has helped faster convergence to an optimal global solution for multiple runs for the test cases considered. For the mutation operation, the upper bound, $UB_{gen,j}$, and the lower bound, $LB_{gen,j}$, of the search space for the $j^\text{th}$ gene in an individual, $x$, in the ${gen}^\text{th}$ generation are given by the equations,
  \begin{equation}
    \label{eq5}
    \alpha  = \alpha_{min}+(1-\frac{gen}{gen_{max}})\times (\alpha_{max}-\alpha_{min}),
  \end{equation}
\begin{equation}
    \label{eq6}
    UB_\alpha  = x_{gen,j} + \alpha \times x_{gen,j}, LB_\alpha  = x_{gen,j} - \alpha \times x_{gen,j},
 \end{equation}
 \begin{equation}
    \label{eq7}
    UB_{gen,j}  = min(UB,UB_\alpha),~\text{and} LB_{gen,j}  = max(LB,LB_\alpha)
 \end{equation}
 where $gen_{max}$ is the maximum number of generations, and $UB$ and $LB$ are the upper and the lower bounds specified for the decision variables in the optimisation problem. A pseudo-code of the MGA is given in Algorithm \ref{alg2}.
 
\subsection{Modified Grey Wolf Optimisation (MGWO)}

 GWO, introduced by Mirjalili \etal, 2014 \cite{48}, is a metaheuristic swarm optimisation algorithm based on the social hierarchy and the hunting mechanism of grey wolves. The standard version of the algorithm is explained in detail in \cite{48}.
 
{\SetAlgoNoLine%
\begin{algorithm}[!t]
\DontPrintSemicolon
 \BlankLine
  \KwInput{$N$, $D$, $max_{ite}$, $max_{count}$}
  \KwOutput{$Alpha$ (Optimal solution)}
  \BlankLine
  Generate initial $N$ search agents, $\vb*X$, using PGF.\;
  \For{$ite\gets1$ \KwTo $max_{ite}$}{
      \For{$i\gets1$ \KwTo $N$}{
    Calculate fitness value, $f(\vb*x_i)$\;
    }
    Find the best three search agents and designate them as $Alpha$, $Beta$, and $Delta$.\;
    \For{$i\gets1$ \KwTo $N$}{
    set $count = 0$.\;
    \While{true}{
    Update position, $\vb*x_i$, of search agent $i$ using equations (\ref{eq8}) to (\ref{eq11}).\;
    Conduct survivability test for the updated search agent, $\vb*x_i$.\;
    \eIf{survivability test passed}{break.}{
    $count=count+1$\;
    \If{$count=max_{count}$}{Set $\vb*x_i$ to the upper bound of the search space.\;
    break.\;}}
    }
    }
    }
\caption{MGWO.}
\label{alg3}
\end{algorithm}}%

 The initial population for the MGWO is generated using the PGF. Consider $N$ as the population size and $D$ as the search space dimension. Let $\vb*x_{i,j}$ denote the position vector of a search agent in the $i^\text{th}$ dimension in the $j^\text{th}$ iteration in the search space. Let the vectors, $\vb*x_\alpha$, $\vb*x_\beta$, and $\vb*x_\delta$ represent the positions of the alpha, beta, and delta in the $j^\text{th}$ iteration. In every iteration, the fitness values of the search agents are calculated, and the positions of alpha, beta, and delta are updated. The positions of all the search agents are updated using the same update equations as those of the standard GWO. The position update equation of a search agent in the $(j+1)^\text{th}$ iteration can be expressed by the equations,
\begin{equation}
\begin{split}
\label{eq8}
    d_{i,\alpha} = |\vb*c_{i,1} \odot \vb*x_{i,\alpha} -\vb*x_{i,j}|,\; d_{i,\beta} = |\vb*c_{i,2} \odot \vb*x_{i,\beta}-\vb*x_{i,j}|,\\ d_{i,\delta} = |\vb*c_{i,3} \odot \vb*x_{i,\delta}-\vb*x_{i,j}|,
    \end{split}
\end{equation}
\begin{equation}
\begin{split}
\label{eq9}
    \vb*x_{1} = \vb*x_{i,\alpha} - d_{i,\alpha}\vb*a_{i,1},\;\vb*x_{2} = \vb*x_{i,\beta}-d_{i,\beta}\vb*a_{i,2}, \\\vb*x_{3} = \vb*x_{i,\delta}-d_{i,\delta}\vb*a_{i,3},~\text{and}
    \end{split}
\end{equation}
\begin{equation}
\label{eq10}
    \vb*x_{i,j+1} = \frac{\vb*x_{1}+\vb*x_{2}+\vb*x_{3}}{3}
\end{equation}
 where $\odot$ represents an element-wise product, $\vb*a_{i,1}$, $\vb*a_{i,2}$, $\vb*a_{i,3}$, $\vb*c_{i,1}$, $\vb*c_{i,2}$, and $\vb*c_{i,3}$ are randomly generated vectors using,
 \begin{equation}
     \label{eq11}
     \vb*a_{d,k} = 2ar_1-a~\text{and}~ \vb*c_{d,k} = 2r_2
 \end{equation}
 where $a$ is linearly decreased from 2 to 0 over the iterations, $r_1$ and $r_2$ are randomly generated numbers from $[0,1]$, $d=1,2,\cdots,D$, and  $k=1,2,3$. After updating the position of each search agent, a survivability test is carried out. If a search agent fails the survivability test, its position is repeatedly updated until a suitable position is obtained or the number of updates reaches a predefined value. In the latter case, the position of the search agent is set to the upper bounds specified by the optimisation problem. The repeated updates help in faster convergence to the optimal solution. This process is carried out for all the search agents in each iteration. At the end of the iteration, the new fitness is calculated, and the alpha, beta, and delta positions are updated. These steps are repeated until the stopping criteria are met. A pseudo-code of the MGWO is given in Algorithm \ref{alg3}.
 
 \subsection{Modified Particle Swarm Optimisation (MPSO)}
 
 Kennedy and Eberhart introduced PSO in 1995\cite{49}. PSO is inspired by the societal behaviour of a swarm of birds. The standard version of the PSO is explained in detail in \cite{50}. 
 In this study, we have used PSO with linearly decreasing inertia weight\cite{51}. In this modified variant of PSO, the initial swarm is generated using the PGF. The position and velocity update equations are the same as the standard PSO.  A survivability test is carried out after each particle's position and velocity updates in every iteration. For the particles failing the test, the velocity and position update is repeated until it passes the test or the number of velocity and position updates reaches a predefined limit. Even after the predefined number of updates, if the particles fail the test, they are replaced by new particles generated by the PGF. The personal best of each particle, the global best, and the PSO parameters are updated. The subsequent iterations are carried out similarly until the stopping criteria are met. Further details and the pseudo-code of the MPSO can be found in \cite{42,43}.
 
\section{Problem Formulation for Op-amp Design}
 The formulation of the optimisation problem for the two op-amp topologies considered in this study is detailed in this section. The methodology used for finding the bounds of the decision variables by the PGF and repair-bounds function for the op-amp optimisation is given in Algorithm \ref{alg4}. Since the study focuses performance evaluation of different algorithms for the area minimisation of the op-amps, we have considered the minimum channel length for the transistors in the respective technologies for the two test cases. All the other specifications for the op-amp design are fixed accordingly.
{\SetAlgoNoLine%
\begin{algorithm}[t]
\DontPrintSemicolon
 \BlankLine
  \KwInput{Op-amp design specifications}
  \KwOutput{Bounds for the decision variables}
  \BlankLine
  Set the bias current bounds from the slew rate and the power specification.\;
  Set the maximum and minimum bounds for $V_{SG}$ of the PMOS load transistors from the $ICMR_{max}$ and the threshold voltage, $V_{TH}$, of the PMOS transistor, respectively. Find the bounds of the widths of the input transistors (M$_\text{3}$ and M$_\text{4}$ for topology 1 as well as topology 2) from the bounds of $V_{SG}$.\;
  Set the minimum and maximum bounds for the widths of the input transistors (M$_\text{3}$ and M$_\text{4}$ for topology 1 as well as topology 2) from the unity gain bandwidth specification, minimum transconductance and $V_{TH}$ of the NMOS transistors.\;
  Set the maximum and minimum bounds for $V_{SG}$ of the tail transistor (M$_\text{5}$ for topology 1 as well as topology 2) from the $ICMR_{min}$ and the threshold voltage, $V_{TH}$, of the NMOS transistor respectively. Find the bounds of the widths of the tail transistor from the bounds of $V_{SG}$.\;
  Find the bounds for the widths of the second stage PMOS transistors ((M$_\text{6}$ for topology 1 and M$_\text{10}$ and M$_\text{11}$ for topology 2) using the maximum output voltage.\;
  Find the bounds for the widths of the second stage NMOS transistors ((M$_\text{7}$ for topology 1 and M$_\text{6}$ to M$_\text{9}$ for topology 2) using the minimum output voltage\;
\caption{To find bounds for the decision variables}
\label{alg4}
\end{algorithm}}%
\subsection{Two-Stage Miller Compensated Op-amp}

  \begin{figure}[!t]
    \centering
    \includegraphics[width=0.8\linewidth]{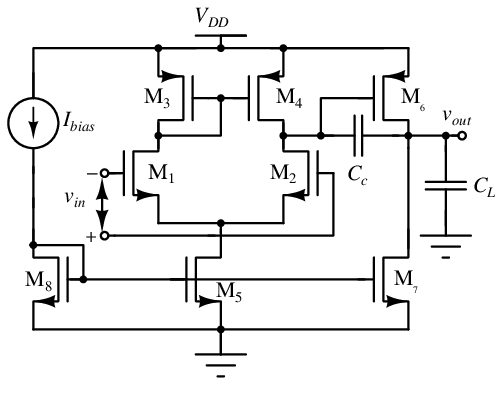}
    \caption{Schematic of a two-stage op-amp with Miller compensation.}
    \label{fig1}
  \end{figure}
  
 A schematic of the two-stage Miller compensated op-amp is shown in Fig.~\ref{fig1}. In Fig.~\ref{fig1}, it is assumed that the transistors M$_\text{1}$ and M$_\text{2}$, M$_\text{3}$ and M$_\text{4}$, and M$_\text{5}$ and M$_\text{8}$ are pairwise matched so that $W_1 = W_2$, $W_3 = W_4$, and $W_5 = W_8$, where $W_k$ is the width of the $k^\text{th}$ transistor. The circuit area is chosen as the fitness function. The widths and the bias current are chosen as the decision variables for the optimisation problem. All the design specifications are considered as constraints. Since all the transistors have to be in the saturation region of operation for the proper working of the circuit, the saturation condition is also added as a constraint.
 
 The position vector, $\vb* x$, and the fitness function, $f(\vb* x)$, are:
  \begin{equation}
    \label{eq21}
    \vb*{x} = [W_{1,2}, W_{3,4}, W_{5,8}, W_6, W_7, I_{bias}]~\text{and}
  \end{equation} 
  \begin{equation}
    \label{eq22}
    f(\vb*{x}) = \Sigma_{i=1}^{M}W_i\times L_i,
  \end{equation}
  respectively, where $M$ is the total number of transistors in the circuit, and $W_i$ and $L_i$ are the width and length of the $i^\text{th}$ transistor. For the circuit in Fig.~\ref{fig1}, $M = 8$. The optimisation problem is framed as:
\begin{equation*}
\begin{aligned}
& \underset{\vb*x}{\text{min}}
&& f(\vb*x) \\
& \text{subject to}
& & \text{Voltage gain } (A_v) \ge\SI{20}{\decibel} \\
&&& \text{Power dissipation }(P) \le\SI{400}{\micro\watt} \\
&&& \text{Slew Rate }(SR) \ge\SI{100}{\volt\per\micro\s} \\
&&& \text{Cut-off frequency }(f_{3dB}) \ge\SI{10}{\mega\hertz} \\
&&& \text{Unity gain bandwidth }(UGB) \ge\SI{100}{\mega\hertz} \\
&&& \text{Phase margin }(PM)\ge\SI{60}{\degree} \\
&&& 2 \le \text{Aspect ratio }(W/L) \le 200 \\
&&& \text{Noise power spectral density }(S_n(f))\\
&&&\le\SI{60}{\nano\volt\per\sqrt{Hz}}\text{ at } \SI{1}{\mega\hertz} \\
&&& \text{All transistors in saturation}.
\end{aligned}
\end{equation*}

 where ${\vb*x}$ and ${f(\vb*x)}$ are given by (\ref{eq21}) and (\ref{eq22}), respectively.

 The two-stage op-amp is designed in a \SI{65}{\nano\meter} technology. The supply voltage is taken as \SI{1.1}{\volt}. The length of all the transistors is fixed to \SI{60}{\nano\meter}, and all the specifications like gain are considered accordingly \cite{52,25}. The value of load capacitor, $C_L$, is taken as \SI{200}{\femto\farad}. To ensure a phase margin of 60$^\circ$, the value of $C_c$ is taken as 0.3 times $C_L$ \cite{53}.
 
\subsection{Folded Cascode Op-amp}
  
\begin{figure}[!t]
    \centering
    \includegraphics[scale=0.8]{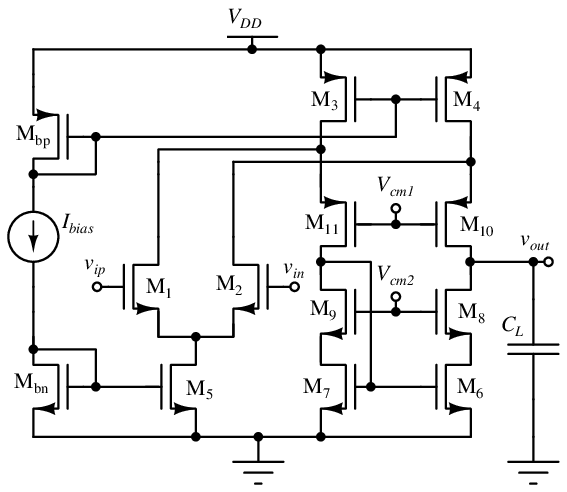}
    \caption{Schematic of a folded cascode op-amp.}
    \label{fig2}
   \end{figure}

 A schematic of the folded cascode op-amp is shown in Fig.~\ref{fig2}. In Fig.~\ref{fig2}, we have assumed that the transistors M$_\text{1}$ and M$_\text{2}$, M$_\text{3}$, M$_\text{4}$, and M$_\text{bp}$, M$_\text{bn}$ and M$_\text{5}$, M$_\text{6}$ and M$_\text{7}$, M$_\text{8}$ and M$_\text{9}$, and M$_\text{10}$ and M$_\text{11}$ are pairwise matched such that $W_1 = W_2$, $W_3 = W_4 = W_{bp}$, $W_{bn} = W_5$, $W_6 = W_7$, $W_8 = W_9$, and $W_{10} = W_{11}$, where $W_k$ is the width of the $k^\text{th}$ transistor. The widths and the bias current of the circuit are selected as the decision variables. The area of the circuit is chosen as the fitness function. The circuit specifications are considered as the constraints along with the saturation condition of the transistors.

 The position vector, $\vb* x$, and the fitness function, $f(\vb* x)$, are:
  \begin{equation}
    \label{eq23}
    \vb*{x} = [W_{1,2}, W_{3,4,bp}, W_{bn,5}, W_{6,7}, W_{8,9}, W_{10,11}, I_{bias}]
  \end{equation} 
  \begin{equation}
    \label{eq24}
    \text{and}~f(\vb*{x}) = \Sigma_{i=1}^{M}W_i\times L_i,
  \end{equation}
  respectively, where $M$ is the number of transistors in the circuit, and $W_i$ and $L_i$ are the width and length of the $i^\text{th}$ transistor. For the circuit in Fig.~\ref{fig2}, $M = 13$. The optimisation problem is framed as:
\begin{equation*}
\begin{aligned}
& \underset{\vb*x}{\text{min}}
& & f(\vb*x) \\
& \text{subject to}
& &\text{Voltage gain }(A_v) \ge\SI{40}{\decibel} \\
&&& \text{Power dissipation }(P) \le\SI{5}{\milli\watt} \\
&&& \text{Slew Rate }(SR)\ge\SI{20}{\volt\per\micro\s} \\
&&& \text{Unity gain bandwidth }(UGB) \ge\SI{40}{\mega\hertz} \\
&&& \text{Phase margin }(PM) \ge\SI{60}{\degree} \\
&&& \frac{4}{3} \le \text{Aspect ratio }(W/L) \le 1000 \\
&&& \text{All transistors in saturation}.
\end{aligned}
\end{equation*}
where ${\vb*x}$ and ${f(\vb*x)}$ are given by (\ref{eq23}) and (\ref{eq24}), respectively.
 
 The folded cascode op-amp is designed in a \SI{180}{\nano\meter} technology. The supply voltage, $V_{DD}$, is taken as \SI{1.8}{\volt}. As in the previous test case, the length of all the transistors is fixed to \SI{180}{\nano\meter}. The load capacitor, $C_L$, is taken as \SI{5}{\pico\farad} \cite{54}.

    \begin{table}[!t]
    \begin{center}
    \caption{Optimum parameters obtained by MABCO, MGA, MGWO, and MPSO for the area optimisation} of two op-amp topologies.
    \label{table2}
    \scalebox{0.85}{
    \begin{tabular}[]{lcccc}
        \hline\hline
        \textbf{Design}&{\bf MABCO}&{\bf MGA}&{\bf MGWO}&{\bf MPSO}\\
        \textbf{Parameter}&&&&\\
        \hline
        \hline
        \multicolumn{5}{c}{\bf Two-stage op-amp}\\
        \hline
        \hline
        $I_{bias}$ (\si{\micro\ampere})&28.8&29.4&29.3&28.6\\
        $W_{1,2}$ (\si{\nano\meter})&259&264&263&257\\
        $W_{3,4}$ (\si{\nano\meter})&797&787&789&801\\
        $W_{5,8}$ (\si{\nano\meter})&121&130&127&121\\
        $W_6$ (\si{\nano\meter})&1114&1099&1104&1113\\
        $W_7$ (\si{\nano\meter})&183&195&191&183\\
        \hline
        \hline
        \multicolumn{5}{c}{\bf Folded cascode op-amp}\\
        \hline
        \hline
        $I_{bias}$ (\si{\micro\ampere})&259.5&257.8&258.0&258.8\\
        $W_{1,2}$ (\si{\nano\meter})&8686&8956&8690&8715\\
        $W_{3,4,bn}$ (\si{\nano\meter})&3417&3450&3367&3418\\
        $W_{5,bp}$ (\si{\nano\meter})&1118&1059&1133&1169\\
        $W_{6,7}$ (\si{\nano\meter})&805&784&801&892\\
        $W_{8,9}$ (\si{\nano\meter})&367&355&367&339\\
        $W_{10,11}$ (\si{\nano\meter})&6156&5945&6220&6035\\
        \hline
        \hline
    \end{tabular}}
    \end{center}
    \end{table}
    
  \begin{table}[!t]
    \caption{Design Specifications obtained for the area optimisation of two op-amp topologies using MABCO, MGA, MGWO, and MPSO in ngspice.}
    \label{table3}
    \begin{center}
    \scalebox{0.8}{
      \begin{tabular}{lccccc}
        \hline
        \hline
        \textbf{Design}&{\bf Specifi-}&{\bf MABCO}&{\bf MGA}&{\bf MGWO}&{\bf MPSO} \\
        \textbf{Criteria}&{\bf cations}&\textbf{}&\textbf{} \\
        \hline
        \hline
        \multicolumn{6}{c}{\bf Two-stage op-amp}\\
        \hline
        \hline
        $A_v$ (\si{\decibel})&$\geq20$&21.9&21.7&21.8&21.9\\
        $f_{3dB}$ (\si{\mega\hertz})&$\geq10$&13.27&13.71&13.61&13.22\\
        $UGB$ (\si{\mega\hertz})&$\geq100$&156.0&157.9&157.6&154.9\\
        $PM$ $(^{\circ})$&$\geq60$&60.0&60.0&60.0&60.0 \\
        $SR$ (\si{\volt\per\micro\second})&$\geq100$&265&269&268&264 \\
        $S_n(f)$@1MHz&$\leq60$&53.30&53.22&53.24&53.35\\ \,\,\,\,(\si{\nano\volt\per\sqrt{Hz}})&&&&&\\
        $P$ (\si{\micro\watt})&$\leq150$&88.2&89.9&89.6&87.5  \\
        Saturation&-&Met&Met&Met&Met\\
        $\textbf{\emph A}$ (\si{\micro\meter^{2}})&$\leq1$&\textbf{0.2191}&\textbf{0.2194}&\textbf{0.2192}&\textbf{0.2192}\\
        \hline
        \hline
        \multicolumn{6}{c}{\bf Folded cascode op-amp}\\
        \hline
        \hline
        $A_v$ (\si{\decibel})&$\geq40$&40.91&40.79&41.03&40.74\\
        $UGB$ (\si{\mega\hertz})&$\geq40$&40.00&40.00&40.00&40.00\\
        $PM$ $(^{\circ})$&$\geq60$&89.87&89.88&89.87&89.88 \\
        $SR$ (\si{\volt\per\micro\second})&$\geq20$&21.22&21.14&21.14&21.14\\
        $P$ (\si{\milli\watt})&$\leq5$&1.233&1.225&1.226&1.230 \\
        Saturation&-&Met&Met&Met&Met\\
        $\textbf{\emph  A}$ (\si{\micro\meter^{2}})&-&\textbf{8.013}&\textbf{8.019}&\textbf{8.014}&\textbf{8.020}\\
        \hline
        \hline
    \end{tabular}}
    \end{center}
    \end{table}

    \begin{table*}[t]
    \begin{center}
    \caption{Simulation results of MABCO, MGA, MGWO, and MPSO for 10 runs for the area optimisation of two-stage op-amp.}
    \label{table1}
    \scalebox{0.9}{
      \begin{tabular}[]{|c|c|c|cccccc|}
      \hline
      {\bf Algorithm}& {\bf Population}&{\bf Iterations}&{\bf Mean} &{\bf Best}&{\bf Worst}&{\bf STDEV}&{\bf MRT}&{\bf CSPR} \\
      &{\bf Size}&&(\si{\micro\meter^{2}})&(\si{\micro\meter^{2}})&(\si{\micro\meter^{2}})&(\si{\micro\meter^{2}})&(\si{\sec})&\\
        \hline
        \hline
        &&100&0.2477&0.2213&0.2671&\num{7.777e-3}&208&1760\\
        &10 & 200&0.2238&0.2191&0.2357&\num{6.338e-3}&474&4590\\
        &&300&0.2198&0.2191&0.2219&\num{8.143e-4}&806&7684\\
        \cline{2-9}
        &&100&0.2259&0.2210&0.2347&\num{4.393e-3}&461&4597\\
        MABCO&20 & 200&\textbf{0.2196}&\textbf{0.2193}&\textbf{0.2206}&\textbf{\num{3.739e-4}}&\textbf{1093}&\textbf{9816}\\
        &&300&0.2194&0.2191&0.2206&\num{4.279e-4}&1722&17027\\
        \cline{2-9}
        &&100&0.2220&0.2200&0.2269&\num{2.209e-3}&832&6939\\
        &30 & 200&0.2195&0.2191&0.2207&\num{5.081e-4}&1999&15675\\
        &&300&0.2194&0.2191&0.2207&\num{5.216e-4}&2962&20717\\
        \hline
        \hline
        &&100&0.2350&0.2240&0.2444&\num{6.911e-3}&181&6086\\
        &10 & 200&0.2295&0.2219&0.2404&\num{5.882e-3}&378&12470\\
        &&300&0.2212&0.2196&0.2228&\num{9.889e-4}&576&19038\\
        \cline{2-9}
        &&100&0.2287&0.2209&0.2412&\num{6.506e-3}&1174&12481\\
        MGA&20 & 200&0.2258&0.2202&0.2348&\num{5.150e-3}&2609&24908\\
        &&300&0.2202&0.2196&0.2210&\num{5.028e-4}&4154&37615\\
        \cline{2-9}
        &&100&0.2293&0.2226&0.2347&\num{5.500e-3}&1610&18766\\
        &30 & 200&0.2243 &0.2206&0.2374&\num{5.272e-3}&3502&37692\\
        &&300&\textbf{0.2202}&\textbf{0.2194}&\textbf{0.2221}&\textbf{\num{8.324e-4}}&\textbf{5497}&\textbf{56686}\\
        \hline
        \hline
        &&100&0.2439&0.2287&0.2553&\num{8.516e-3}&1425&4436\\
        &10 & 200&0.2245&0.2227&0.2266&\num{1.545e-5}&3398&9644\\
        &&300&\textbf{0.2196}&\textbf{0.2194}&\textbf{0.2198}&\textbf{\num{1.193e-4}}&\textbf{4747}&\textbf{12623}\\
        \cline{2-9}
        &&100&0.2413&0.2346&0.2530&\num{6.205e-3}&4156&8662\\
        MGWO&20 & 200 &0.2243&0.2217&0.2258&\num{1.228e-3}&8965&17751\\
        &&300&0.2194&0.2192&0.2198&\num{1.716e-4}&12726&25654\\
        \cline{2-9}
        &&100&0.2393&0.2324&0.2488&\num{5.409e-3}&7350&16439\\
        &30 & 200&0.2232&0.2207&0.2251&\num{1.251e-3}&15577&33239\\
        &&300&0.2194&0.2192&0.2195&\num{1.271e-4}&22180&46868\\
        \hline
        \hline
        &&100&0.2259&0.2219&0.2323&\num{3.708e-3}&343&6266\\
        &10 & 200&0.2196 &0.2192&0.2203&\num{3.773e-4}&1085&13231\\
        &&300&0.2194&0.2192&0.2198&\num{2.055e-4}&1780&19245\\
        \cline{2-9}
        &&100&0.2226&0.2203&0.2247&\num{1.446e-3}&1432&17838\\
        MPSO&20 & 200&\textbf{0.2194}&\textbf{0.2192}&\textbf{0.2195}&\textbf{\num{8.020e-5}}&\textbf{4761}&\textbf{32456}\\
        &&300&0.2194&0.2192&0.2195&\num{7.183e-5}&7320&41786\\
        \cline{2-9}
        &&100&0.2217&0.2209&0.2242&\num{1.087e-3}&3180&31451\\
        &30 & 200&0.2197&0.2192&0.2213&\num{6.355e-4}&9515&55671\\
        &&300&0.2196&0.2192&0.2213&\num{6.419e-4}&14665&76589\\
        \hline
      \end{tabular} }
    \end{center}
    \end{table*}

       \begin{table*}[t]
    \begin{center}
    \caption{Simulation results of MABCO, MGA, MGWO, and MPSO for 10 runs for the area optimisation of folded cascode op-amp.}
    \label{table4}
    \scalebox{0.9}{
      \begin{tabular}[]{|c|c|c|cccccc|}
      \hline
      {\bf Algorithm}& {\bf Population}&{\bf Iterations}&{\bf Mean} &{\bf Best}&{\bf Worst}&{\bf STDEV}&{\bf MRT}&{\bf CSPR}\\
      &{\bf Size}&&(\si{\micro\meter^{2}})&(\si{\micro\meter^{2}})&(\si{\micro\meter^{2}})&(\si{\micro\meter^{2}})&(\si{\sec})&\\
        \hline
        \hline
        &&100&9.628&8.118&18.10&\num{3.083e0}&358&15398\\
        &10 & 200&8.614&8.057&12.094&\num{1.240e0}&501&26656\\
        &&300&8.155&8.030&8.645&\num{1.814e-1}&653&43642\\
        \cline{2-9}
        &&100&8.146&8.044&8.431&\num{1.174e-1}&422&26105\\
        MABCO&20 & 200&8.039&8.016&8.137&\num{3.528e-2}&730&33530\\
        &&300&8.026&8.014&8.067&\num{1.606e-2}&984&47736\\
        \cline{2-9}
        &&100&8.073&8.023&8.190&\num{5.912e-2}&582&40895\\
        &30 & 200&\textbf{8.024} &\textbf{8.013}&\textbf{8.065}&\textbf{\num{1.507e-2}}&\textbf{877}&\textbf{67485}\\
        &&300&8.015&8.013&8.019&\num{1.882e-3}&1193&89968\\
        \hline
        \hline
        &&100&8.257&8.038&8.463&\num{1.197e-1}&165&11406\\
        &10 & 200&8.1234&8.038&8.200&\num{6.430e-2}&220&17117\\
        &&300&8.064&8.020&8.166&\num{4.975e-2}&266&21977\\
        \cline{2-9}
        &&100&8.325&8.090&9.190&\num{3.217e-1}&236&27628\\
        MGA&20 & 200&8.115&8.042&8.208&\num{6.421e-2}&315&38574\\
        &&300&8.032&8.023&8.043&\num{6.429e-3}&379&46711\\
        \cline{2-9}
        &&100&8.2263&8.036&8.423&\num{1.352e-1}&314&39252\\
        &30 & 200&8.144 &8.024&8.282&\num{9.443e-2}&419&54643\\
        &&300&\textbf{8.031}&\textbf{8.019}&\textbf{8.041}&\textbf{\num{7.525e-3}}&\textbf{443}&\textbf{67885}\\
        \hline
        \hline
        &&100&8.308&8.171&18.611&\num{1.362e-1}&228&11979\\
        &10 & 200&8.082&8.051&8.182&\num{3.730e-2}&292&20224\\
        &&300&8.026&8.014&8.054&\num{1.276e-2}&334&24060\\
        \cline{2-9}
        &&100&8.208&8.107&8.302&\num{6.636e-2}&378&27387\\
        MGWO&20 & 200 &8.050&8.042&8.061&\num{7.393e-3}&501&41719\\
        &&300&\textbf{8.019}&\textbf{8.016}&\textbf{8.025}&\textbf{\num{2.987e-3}}&\textbf{565}&\textbf{50073}\\
        \cline{2-9}
        &&100&8.218&8.135&8.296&\num{5.183e-2}&558&46168\\
        &30 & 200&8.046 &8.033&8.090&\num{1.611e-2}&724&70241\\
        &&300&8.019&8.014&8.026&\num{4.924e-3}&813&83283\\
        \hline
        \hline
        &&100&8.706&8.246&9.099&\num{2.834e-1}&154&4172\\
        &10 & 200&8.250&8.142&8.394&\num{8.937e-2}&204&7459\\
        &&300&8.115&8.066&8.169&\num{3.057e-2}&236&9000\\
        \cline{2-9}
        &&100&8.357&8.185&8.542&\num{1.089e-1}&293&8307\\
        MPSO&20 & 200&8.149 &8.057&8.248&\num{5.332e-2}&358&16306\\
        &&300&8.098&8.057&8.133&\num{2.564e-2}&408&22296\\
        \cline{2-9}
        &&100&8.214&8.138&8.247&\num{3.517e-2}&368&13367\\
        &30 & 200&8.086 &8.038&8.133&\num{3.082e-2}&440&24988\\
        &&300&\textbf{8.047}&\textbf{8.020}&\textbf{8.098}&\textbf{\num{2.366e-2}}&\textbf{496}&\textbf{34590}\\
        \hline
      \end{tabular} }
    \end{center}
    \end{table*}

    \begin{table}[!t]
    \caption{Comparison of standard versions with modified versions of the algorithms for 10 runs for the area optimisation of two-stage Miller compensated op-amp in \SI{65}{\nano\meter}.}
    \begin{center}
      \label{table5}
      \centering
      \begin{tabular}[!t]{lcccc}
        \hline
        \hline
        \bf Algorithm & {\bf Best}&{\bf Worst}& {\bf Mean}& {\bf STDEV}\\
        &(\si{\micro\meter^{2}})&(\si{\micro\meter^{2}})&(\si{\micro\meter^{2}})&(\si{\micro\meter^{2}})\\
        \hline
        \hline
        SABCO &0.3010&0.3750&0.3230&0.02\\
        MABCO &0.2213&0.2671&0.2477&0.008\\
        SGA &0.3120&0.4099&0.3428&0.05\\
        MGA &0.2240&0.2444&0.2350&0.007\\
        SGWO &0.4081&0.5968&0.4951&0.07\\
        MGWO &0.2287&0.2553&0.2439&0.009\\
        \hline
        \hline
      \end{tabular}
    \end{center}
  \end{table}

\section{Simulation Results and Discussion}

 The performance of the four EAs, namely, MABCO, MGA, MGWO, and MPSO, were compared in terms of convergence time and accuracy for different population sizes and iterations. The simulations were carried out on a desktop with Intel\textsuperscript{\textregistered} Core\textsuperscript{TM} i7-9700F CPU @3.00\,GHz and 16\,GB RAM. Implementing parallel processing for the update of each solution in every iteration reduced the execution time substantially for all the algorithms. All the algorithms were coded in Python, and the circuit simulations were performed in ngspice. Population sizes of 10, 20, and 30 were considered for each algorithm, with the maximum number of iterations set to 300 for each run. A statistical study is needed to validate the robustness of the proposed methodology with the different algorithms. For this, ten simulation runs of each algorithm starting from ten different initialisations were carried out, as in \cite{52}. For all algorithms, the best value, the worst value, the mean value, and the standard deviation (STDEV) of the fitness function value, along with the mean run time (MRT) and the number of circuit simulations per run (CSPR), were recorded for the ${\text{100}}^{\text{th}}$, ${\text{200}}^{\text{th}}$, and the ${\text{300}}^{\text{th}}$ iterations, respectively. The results for the area optimisation of the two-stage and the folded cascode op-amps for the different algorithms are summarized in Table.~\ref{table1} and Table.~\ref{table4}. The design parameters and the corresponding specifications of the best solution obtained by the algorithms for a population size of 30 for the two op-amp topologies are given in Table.~\ref{table2} and Table.~\ref{table3}, respectively. The introduction of PGF and the repair-bounds function in the two-stage op-amp test case resulted in an average reduction of 69\% in the CSPR for every optimisation run. Similarly, for the folded cascode op-amp test case, there was a 43\% reduction in CSPR for each run.
   
 \begin{figure*}[!t]
  \centering
  \begin{tabular}{ccc}
   \includegraphics[scale=0.15]{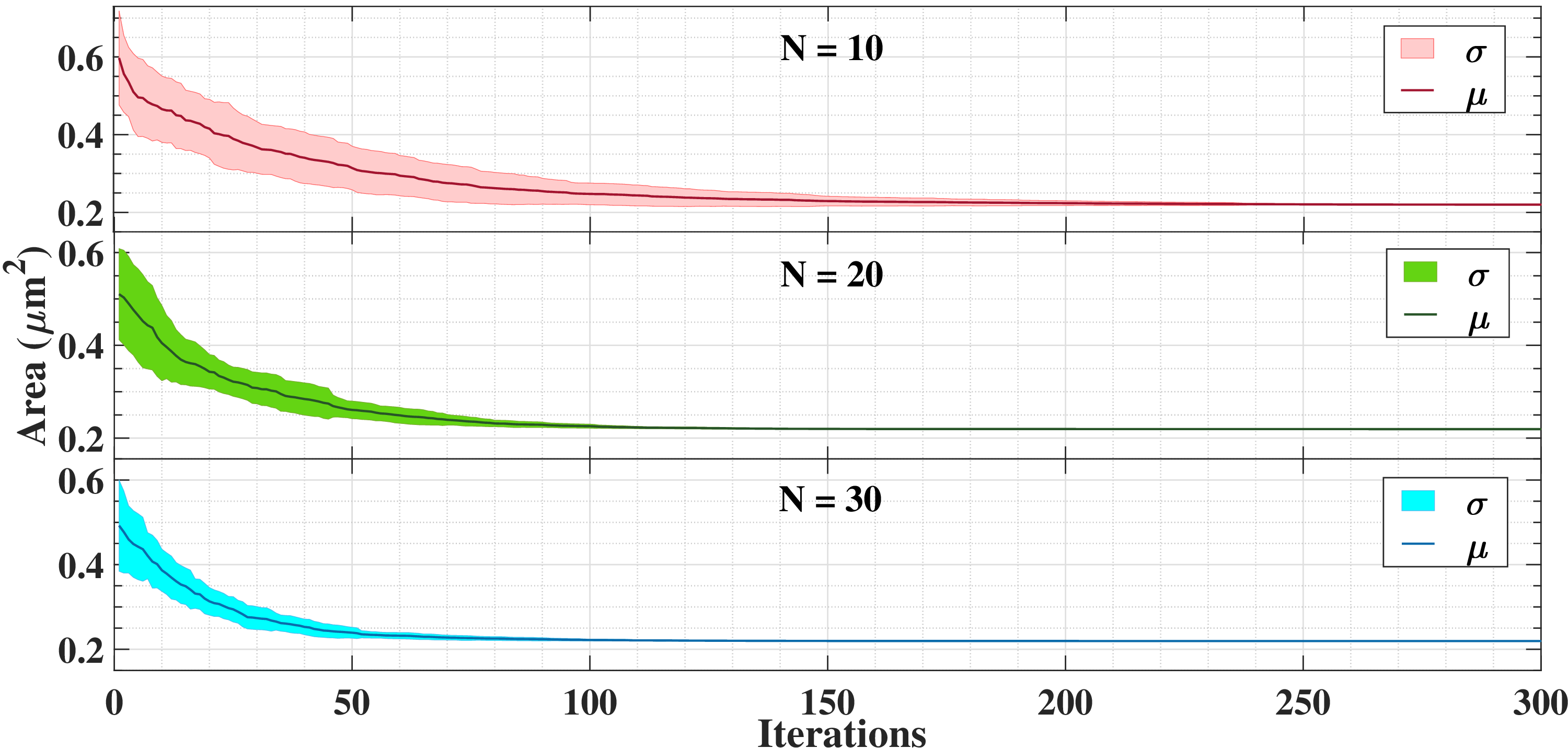} & & \includegraphics[scale=0.15]{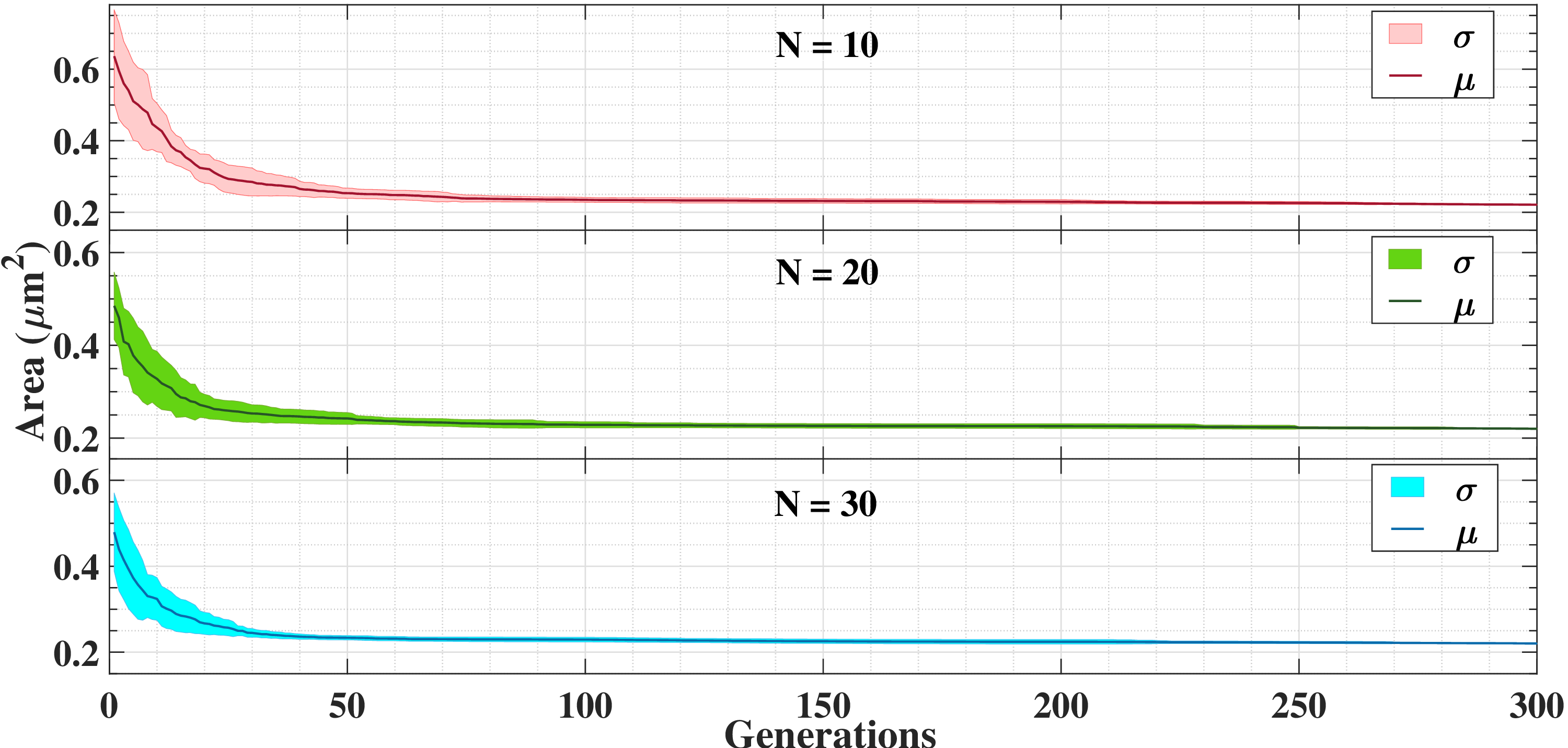} \\
   (a) & & (b) \\[6pt]
   \includegraphics[scale=0.15]{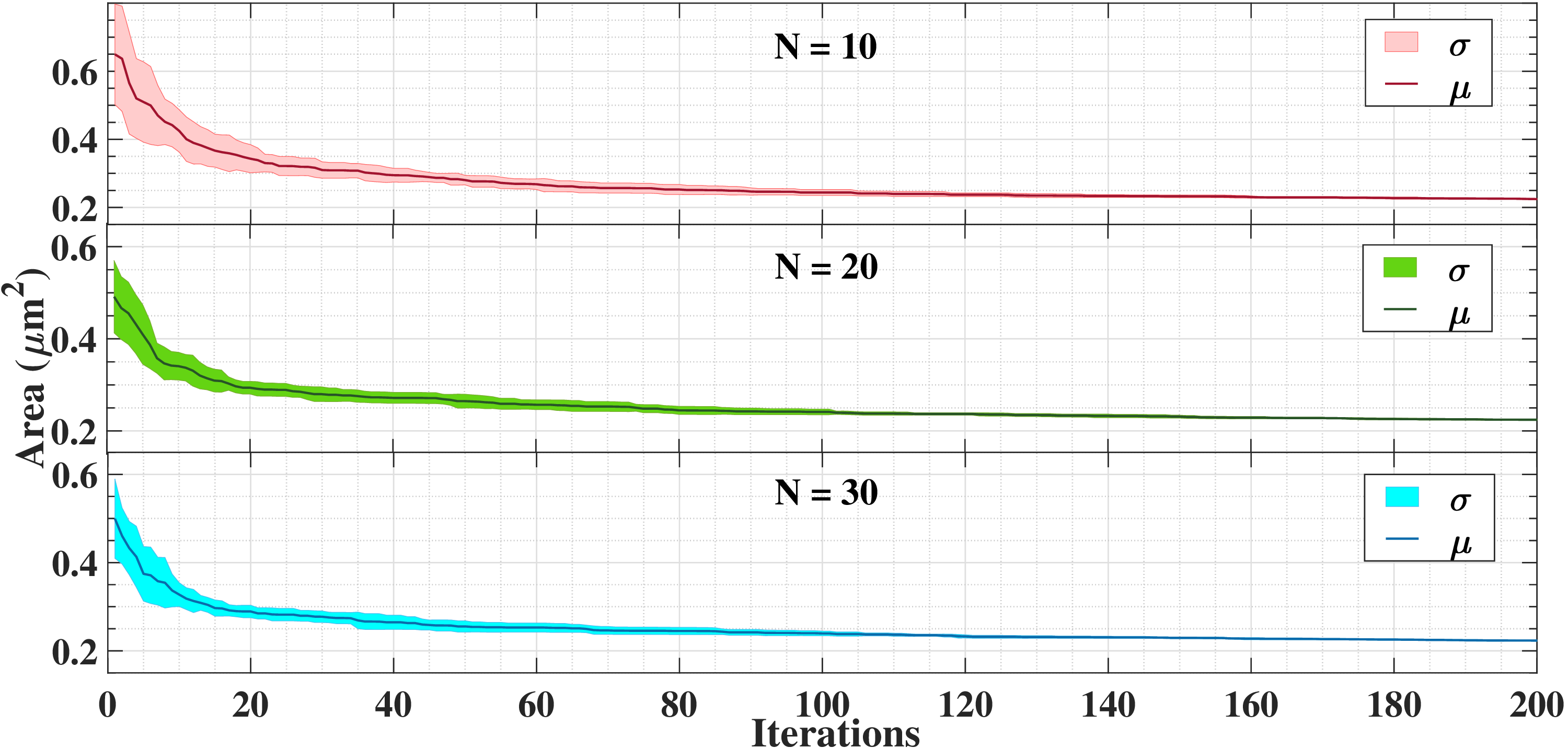} & & \includegraphics[scale=0.15]{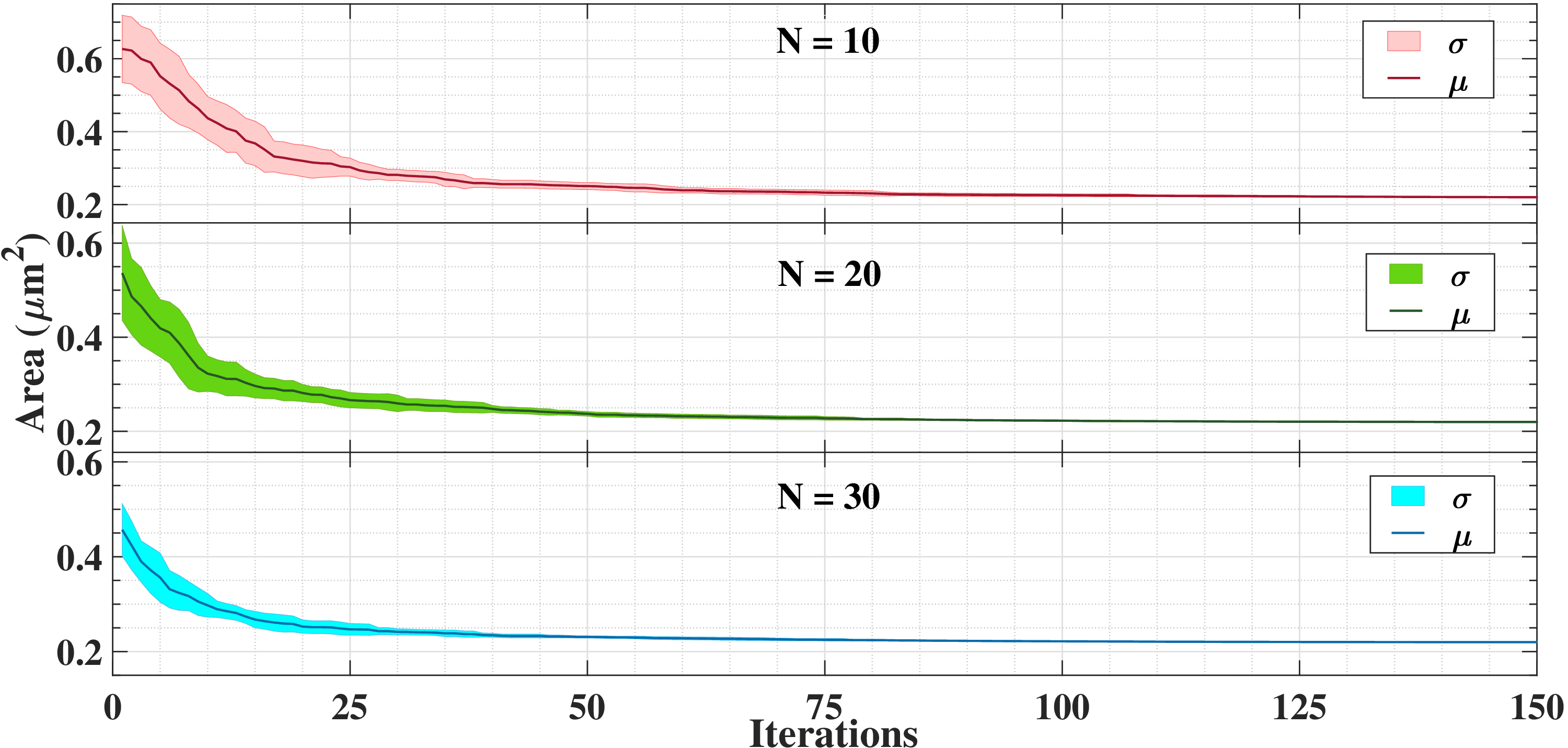} \\
   (c) & & (d) 
  \end{tabular}
  \caption{Convergence plots of (a) MABCO, (b) MGA, (c) MGWO, and (d) MPSO for two-stage op-amp with population sizes of 10, 20, and 30.}
  \label{fig3}
 \end{figure*}
    
 \begin{figure*}[!t]
  \centering
  \begin{tabular}{ccc}
   \includegraphics[scale=0.15]{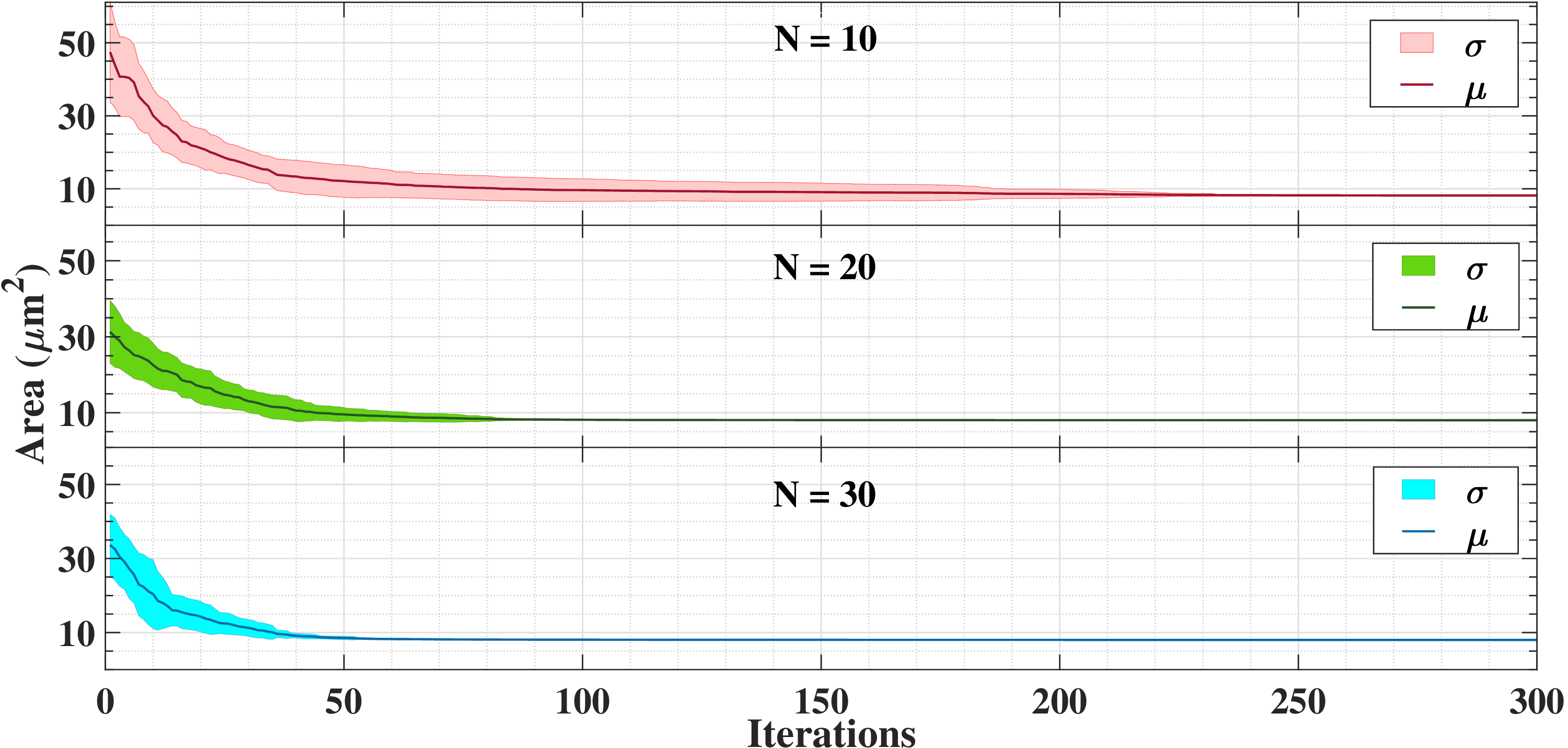} & & \includegraphics[scale=0.15]{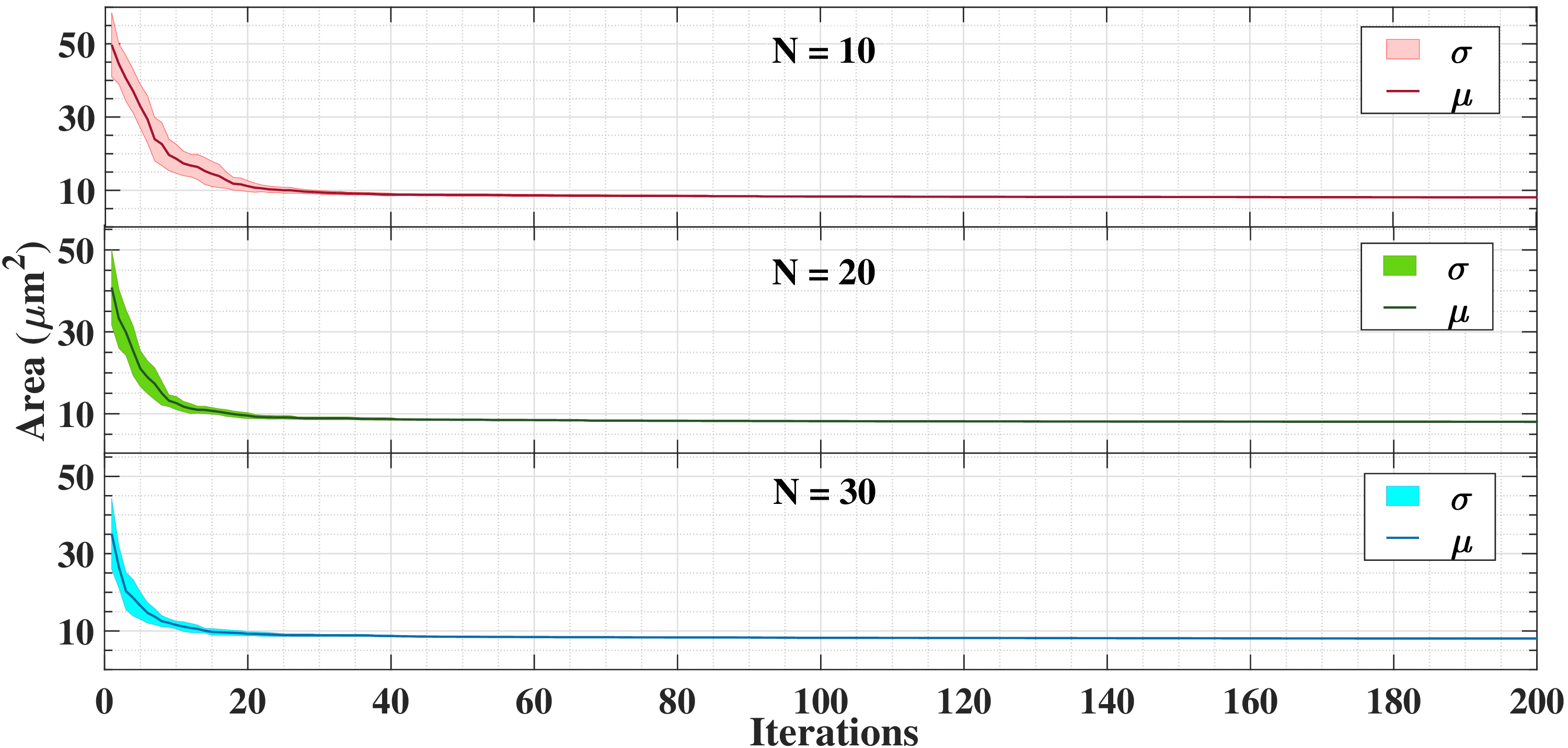} \\
   (a) & & (b) \\[6pt]
   \includegraphics[scale=0.15]{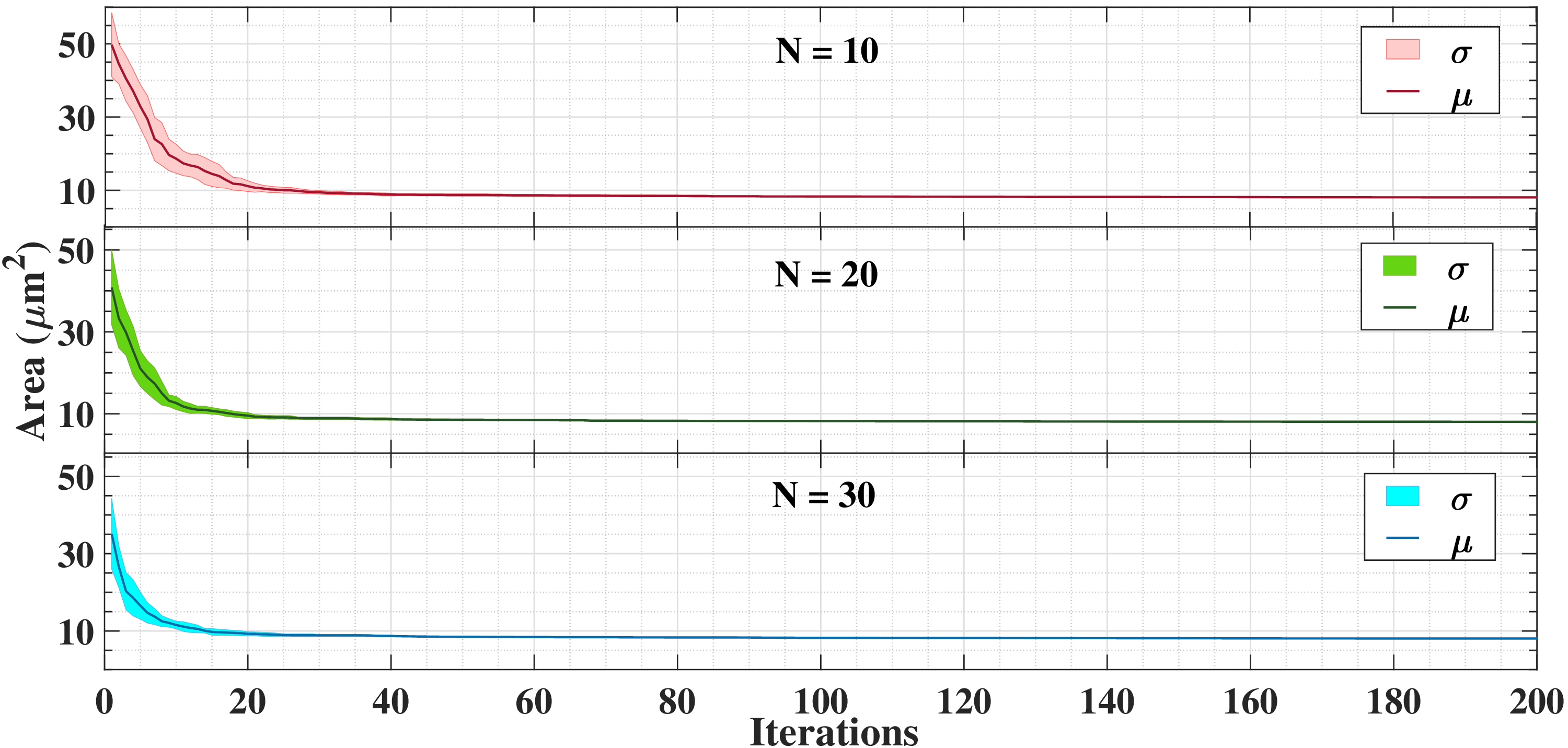} & & \includegraphics[scale=0.15]{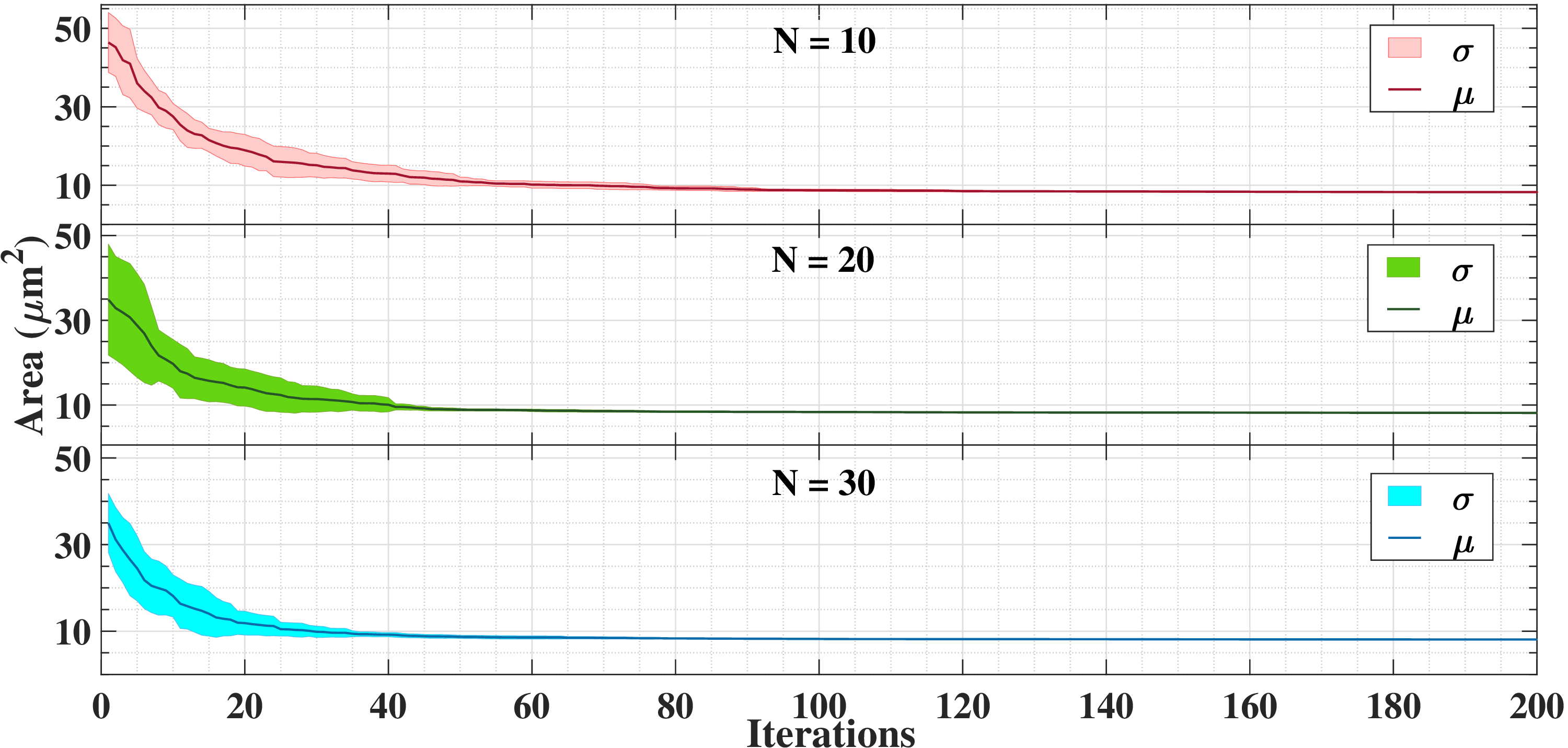} \\
   (c) & & (d) 
  \end{tabular}
  \caption{Convergence plots of (a) MABCO, (b) MGA, (c) MGWO, and (d) MPSO for folded cascode op-amp with population sizes of 10, 20, and 30.}
  \label{fig7}
 \end{figure*}

  \begin{figure*}[!t]
  \centering
  \begin{tabular}{ccc}
   \includegraphics[scale=0.165]{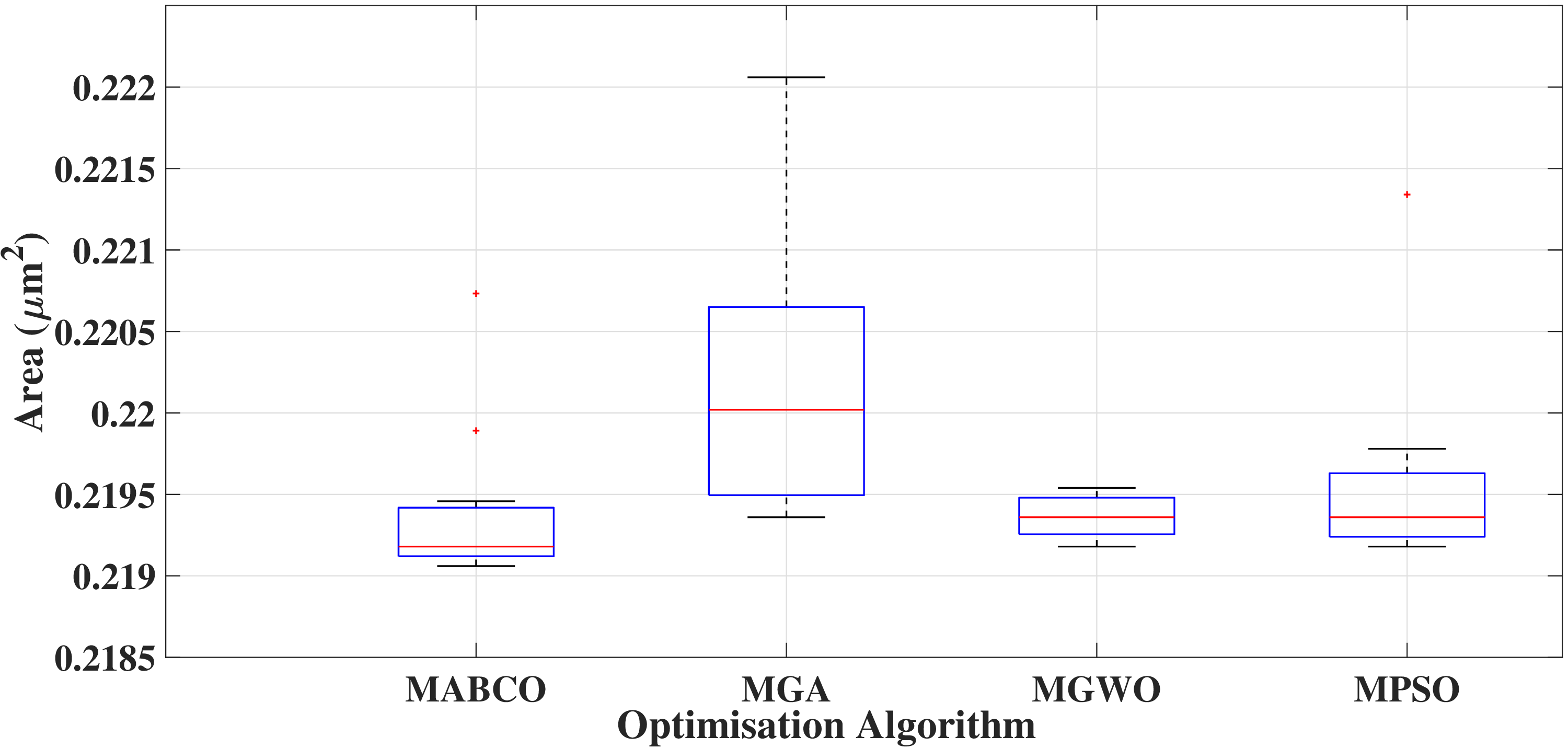} & & \includegraphics[scale=0.165]{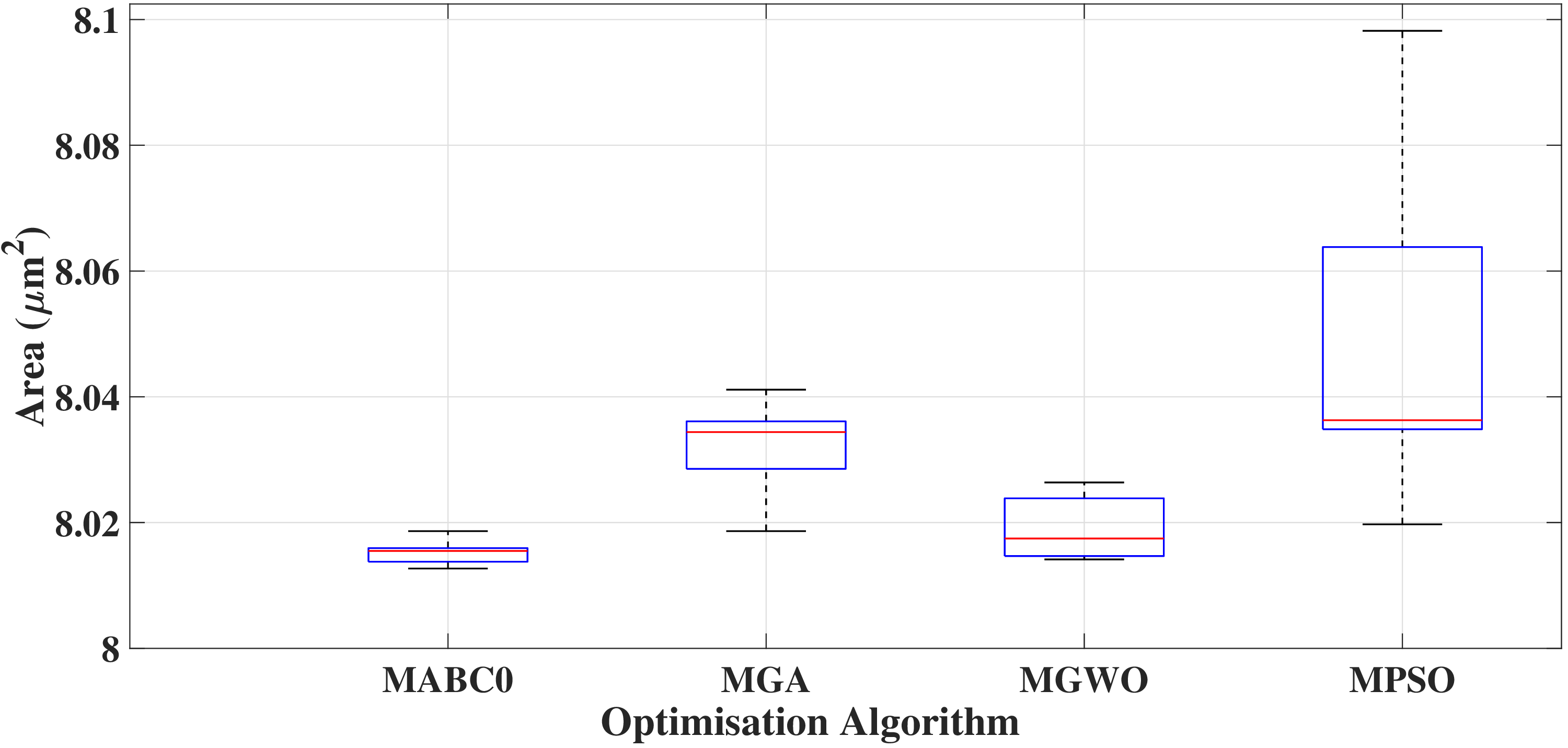} \\
   (a) & & (b) \\
  \end{tabular}
  \caption{Box and whisker plots for area optimisation of (a) two-stage op-amp and (b) folded cascode op-amp for a population size of 30.}
  \label{fig11}
 \end{figure*}

    \begin{table}[!t]
    \caption{Comparison with Bayesian Optimisation for the unconstrained optimisation of two-stage op-amp in \SI{180}{\nano\meter}.}
    \begin{center}
      \label{table6}
      \centering
      \begin{tabular}[!t]{lcccc}
        \hline
        \hline
        \bf Algorithm & {\bf Best}&{\bf Worst}& {\bf Mean}& {\bf STDEV}\\
        \hline
        \hline
        MACE \cite{17}&690.36&690.27&690.34&0.03\\
        MPSO \cite{43}&2110.1&2100.1&2104.8&2.9\\
        MABCO &2112.8&2106.7&2108.8&1.4\\
        MGA &2090.1&2077.1&2088.2&3.2\\
        MGWO &2108.1&2099.1&2103.8&2.8\\
        \hline
        \hline
      \end{tabular}
      \\\scriptsize{$^\ast$ Not with technology file from the same foundry}
    \end{center}
  \end{table}

    \begin{table}[!t]
    \caption{Comparison with flower pollination algorithm \cite{55} for the power optimisation of two-stage op-amp in \SI{180}{\nano\meter}.}
    \label{table7}
    \begin{center}
    \scalebox{0.7}{
      \begin{tabular}{lcccccc}
        \hline
        \hline
        \textbf{Design}&{\bf Specifi-}&{\bf FPA}&{\bf MABCO}&{\bf MGA}&{\bf MGWO}&{\bf MPSO} \\
        \textbf{Criteria}&{\bf cations}&\cite{55}&\textbf{}&\textbf{} \\
        \hline
        \hline
        $A_v$ (\si{\decibel})&$\geq65$&69.8&67.83&66.65&66.28&65.52\\
        $UGB$ (\si{\mega\hertz})&$\geq5$&5.252&8.025&9.694&10.873&9.281\\
        $PM$ $(^{\circ})$&$\geq60$&66.45&63.43&60.16&61.09&60.22 \\
        $SR$ (\si{\volt\per\micro\second})&$\geq10$&10.38&10.22&10.03&10.04&10.68\\
        $CMRR$ (\si{\decibel})&$\geq80$&88&80.03&80.97&80.50&80.01\\
        $C_L$ (\si{\pico\farad})&$\geq10$&10&10&10&10&10\\
        $A$ (\si{\micro\meter^{2}})&$\leq300$&252&191&166&168&187\\
        $\textbf{P}$ (\si{\micro\watt})&$\leq3000$&\textbf{712.2}&\textbf{317.9}&\textbf{361.5}&\textbf{354.9}&\textbf{332.3}  \\
        \hline
        \hline
        \end{tabular}}
        \\\scriptsize{$^\ast$ Not with the technology file from the same foundry}
    \end{center}
    \end{table}

 For MABCO, the bounds for the limit parameter value were set as $limit_{min}$ = 5 and $limit_{max}$ = 15. Convergence plots showing the mean and STDEV of the fitness function value for multiple runs for different swarm sizes for the two test cases are given in Fig.~\ref{fig3}(a) and Fig.~\ref{fig7}(a), respectively. The figures show that as the population size increases, the algorithm converges to the optimal solution in fewer iterations. For the test case of the two-stage op-amp in Table \ref{table1}, it is inferred that as the population size is increased from 10 to 20, the mean and the STDEV of the fitness function value obtained for multiple runs decrease. Further, when the population size is increased from 20 to 30, there is not much variation in the mean and the STDEV values of the results. For a swarm size of 10, it is seen that the convergence to the optimal solution occurs after the ${\text{250}}^\text{th}$ iteration. For the swarm size of 20 and 30, the convergence to the optimal solution happens before the ${\text{200}}^\text{th}$ iteration. As the population size increases, the MRT and CSPR for each simulation increases. Considering all the above factors, a swarm size of 20 with maximum iterations of 200 gives the best result for this test case. Similarly, for the test case of folded cascode op-amp in Table \ref{table4}, a population size of 30 with maximum iterations of 200 is found to be the best choice in the same manner.

 For MGA, the convergence plots for multiple runs for the area optimisation of two-stage op-amp and folded cascode op-amp are given in Fig.~\ref{fig3}(b) and Fig.~\ref{fig7}(b), respectively. For the folded cascode op-amp test case, it is seen that as the population size is increased, the algorithm converges to the optimal solution in a fewer number of iterations. The STDEV of the results for multiple runs for a population size of 10 is larger when compared to the results with population sizes of 20 and 30. With a population size of 30, the mean, STDEV, and the best value obtained for the fitness function are better than the results obtained with population sizes of 10 and 20. For the considered population sizes, convergence to the optimal solution occurs between the ${\text{200}}^\text{th}$ and the ${\text{300}}^\text{th}$ generations. Hence, a population size of 30 with a maximum generation of 300 is found to give the best results for the folded cascode op-amp optimisation problem. For the test case of the two-stage op-amp, a population size of 30 with a generation of 300 is similarly found to give the best results.
 
 For MGWO, the convergence plots obtained for the area optimisation of the two-stage op-amp and folded cascode op-amp are given in Fig.~\ref{fig3}(c) and Fig.~\ref{fig7}(c), respectively. For the two-stage op-amp test case, the mean, STDEV, and the best result obtained for the population sizes of 10, 20, and 30  using MGWO are comparable. The best results are obtained with the maximum iterations as 300 for all the considered population sizes. As the number of iterations is increased, it is found that the STDEV, mean, and the best value of the results have improved for all the considered population sizes. After considering the MRT and CSPR as well, a population size of 10 with the number of iterations as 300 is found to give the best result for MGWO for the two-stage op-amp test case. Similarly, for the folded cascode op-amp test case, a population size of 20 with maximum iterations of 300 is found to be the best choice.
 
 For MPSO, the inertia weight, $w$, was varied linearly between $w_{max}= 0.8$ and $w_{min}= 0.5$ for the two-stage op-amp test case. For the optimisation of folded cascode op-amp, $w_{max}$ and $w_{min}$ were chosen as 0.8 and 0.3, respectively. The parameters, $c_1$ and $c_2$, were set to 1.7 for both the test cases. The fine-tuning of PSO parameters for the two cases considered was found to be more time-consuming. The mean and STDEV curves for the results for the two-stage op-amp and the folded cascode op-amp are given in Fig.~\ref{fig3}(d) and Fig.~\ref{fig7}(d), respectively. For the test case of the two-stage op-amp, for the population sizes of 10, 20, and 30,  there is no significant change in results between the ${\text{200}}^\text{th}$ and the ${\text{300}}^\text{th}$ iteration. The results obtained with a population size of 20 are better than those with a population size of 10. There is no significant variation in results when the population size increases from 20 to 30. After considering the MRT and CSPR, a population size of 20 with maximum iterations of 200 is found to give the best results for this test case. For the folded cascode op-amp, a maximum iteration of 300 gives the best results for all the population sizes. A population size of 30 gives better results than population sizes of 10 and 20. For this test case, a population size of 30 with maximum iterations of 300 is found to be the best choice.

 \par Table.~\ref{table5} shows the results of the area optimisation of two-stage Miller compensated op-amp in \SI{65}{\nano\meter} for the standard as well as modified versions of the algorithms. The best, worst, mean, and standard deviation of the results for 10 consecutive runs with a population size of 10 and maximum iterations of 100 is given. SABCO, SGA, and SGWO correspond to the standard versions of the algorithms. It can be seen that the modified algorithms converge to better values for the objective function with considerably less standard deviation for multiple runs. The performance of the proposed methodology with MPSO for the area optimisation of a two-stage op-amp was compared with other reported studies in \cite{43}, and the results clearly showed that the algorithm worked better for all the test cases considered. It is to be noted that the MPSO in \cite{43} was implemented in MATLAB, and the simulations were run serially. All the algorithms in this study have better performances than the MPSO reported in \cite{43}. The proposed method with the modified algorithms has been compared with \cite{17} for the unconstrained optimisation of the op-amp topology reported, and the results are tabulated in Table.~\ref{table6}. The results clearly show that the proposed methodology is able to obtain a better result in terms of the best optimal value obtained. The proposed methodology has also been compared with \cite{55} for the power optimisation of a two-stage Miller compensated op-amp in \SI{180}{\nano\meter}, and the results are tabulated in Table.~\ref{table7}. All four algorithms are able to achieve a better value for power than the value reported in \cite{55}. 

    \begin{table}[!t]
    \begin{center}
    \caption{Optimum parameters obtained by MABCO, MGA, MGWO, and MPSO for the noise optimisation of two op-amp topologies.}
    \label{table8}
    \scalebox{0.85}{
    \begin{tabular}[]{lcccc}
        \hline\hline
        \textbf{Design}&{\bf MABCO}&{\bf MGA}&{\bf MGWO}&{\bf MPSO}\\
        \textbf{Parameter}&&&&\\
        \hline
        \hline
        \multicolumn{5}{c}{\bf Two-stage op-amp}\\
        \hline
        \hline
        $I_{bias}$ (\si{\micro\ampere})&90.3&90.7&90.9&90.8\\
        $W_{1,2}$ (\si{\micro\meter})&0.74&0.74&0.74&0.74\\
        $W_{3,4}$ (\si{\micro\meter})&3.33&3.32&3.31&3.33\\
        $W_{5,8}$ (\si{\micro\meter})&0.80&0.91&0.74&0.95\\
        $W_6$ (\si{\micro\meter})&6.05&5.66&5.92&5.45\\
        $W_7$ (\si{\micro\meter})&0.93&1.06&0.96&1.15\\
        \hline
        \hline
        \multicolumn{5}{c}{\bf Folded cascode op-amp}\\
        \hline
        \hline
        $I_{bias}$ (\si{\micro\ampere})&386.6&380.8&414.8&393.1\\
        $W_{1,2}$ (\si{\micro\meter})&49.88&49.75&49.05&49.86\\
        $W_{3,4,bn}$ (\si{\micro\meter})&4.19&4.39&4.82&4.40\\
        $W_{5,bp}$ (\si{\micro\meter})&6.58&9.54&8.04&7.65\\
        $W_{6,7}$ (\si{\micro\meter})&6.37&5.79&6.07&6.21\\
        $W_{8,9}$ (\si{\micro\meter})&0.416&0.396&0.465&0.434\\
        $W_{10,11}$ (\si{\micro\meter})&13.77&11.18&12.29&12.57\\
        \hline
        \hline
    \end{tabular}}
    \end{center}
    \end{table}
    
  \begin{table}[!t]
    \caption{Design Specifications obtained for the noise optimisation of two op-amp topologies using MABCO, MGA, MGWO, and MPSO in ngspice.}
    \label{table9}
    \begin{center}
    \scalebox{0.8}{
      \begin{tabular}{lccccc}
        \hline
        \hline
        \textbf{Design}&{\bf Specifi-}&{\bf MABCO}&{\bf MGA}&{\bf MGWO}&{\bf MPSO} \\
        \textbf{Criteria}&{\bf cations}&\textbf{}&\textbf{} \\
        \hline
        \hline
        \multicolumn{6}{c}{\bf Two-stage op-amp}\\
        \hline
        \hline
        $A_v$ (\si{\decibel})&$\geq20$&23.6&23.2&23.3&22.9\\
        $f_{3dB}$ (\si{\mega\hertz})&$\geq10$&35.91&37.33&37.15&38.65\\
        $UGB$ (\si{\mega\hertz})&$\geq100$&507.8&504.9&514.4&505.3\\
        $PM$ $(^{\circ})$&$\geq60$&60.2&60.0&60.9&60.3 \\
        $SR$ (\si{\volt\per\micro\second})&$\geq500$&721&716&702&697\\
        $A$ (\si{\micro\meter^{2}})&$\leq1$&0.999&0.999&0.988&0.998\\
        $P$ (\si{\micro\watt})&$\leq500$&260&258&266&259  \\
        Saturation&-&Met&Met&Met&Met\\
        $\textbf{\emph Noise}$@1MHz&&\textbf{25.95}&\textbf{25.89}&\textbf{25.92}&\textbf{25.85}\\ \,\,\,\,(\si{\nano\volt\per\sqrt{Hz}})&&&&&\\
        \hline
        \hline
        \multicolumn{6}{c}{\bf Folded cascode op-amp}\\
        \hline
        \hline
        $A_v$ (\si{\decibel})&$\geq40$&40.0&40.0&40.1&40.0\\
        $UGB$ (\si{\mega\hertz})&$\geq40$&90.74&87.88&93.81&90.10\\
        $PM$ $(^{\circ})$&$\geq60$&87.6&87.7&87.66&87.6 \\
        $SR$ (\si{\volt\per\micro\second})&$\geq20$&36.2&35.9&38.2&36.2\\
        $P$ (\si{\milli\watt})&$\leq2$&1.84&1.82&1.98&1.88 \\
        Saturation&-&Met&Met&Met&Met\\
        $A$ (\si{\micro\meter^{2}})&$\leq30$&29.99&29.97&29.93&29.99\\
        $\textbf{\emph Noise}$@10kHz&-&\textbf{45.71}&\textbf{46.10}&\textbf{45.79}&\textbf{45.67}\\ \,\,\,\,(\si{\nano\volt\per\sqrt{Hz}})&&&&&\\
        \hline
        \hline
    \end{tabular}}
    \end{center}
    \end{table}

 Noise optimisation has also been carried out for the two op-amp topologies considered. The design parameters and the corresponding specifications of the best solution obtained by the algorithms for 20 consecutive runs for a population size of 20 for the two op-amp topologies are given in Table.~\ref{table8} and Table.~\ref{table9}, respectively. It is seen from the results that all four algorithms are converging to a narrow range. This shows that the proposed methodology can be used for the optimisation of any desired circuit specification.
 
\section{Conclusion}    

 Four different evolutionary algorithms, namely, artificial bee colony optimisation, genetic algorithm, grey wolf optimisation, and particle swarm optimisation, have been used to design a two-stage op-amp in \SI{65}{\nano\meter} technology and a folded cascode op-amp in \SI{180}{\nano\meter} technology in this study. All the algorithms are modified for optimal performance. The proposed optimisation approach has used a particle generation function, and repair-bounds function for targeted search in the search space, resulting in fewer spice calls and faster convergence for all the algorithms. A performance evaluation of all the modified algorithms is conducted based on population size, convergence, and run-time by performing exhaustive simulations. The optimal results obtained by the algorithms are found to meet all the required circuit specifications. The implementation of parallel computation has reduced the run time considerably for all the algorithms. The box and whisker plot of the fitness function value for ten consecutive runs for a population size of 30 for the two op-amp topologies are given in Fig.~\ref{fig11}(a) and Fig.~\ref{fig11}(b), respectively. For the two-stage op-amp test case in Fig.~\ref{fig11}(a), it is clear that MABC, MGWO, and MPSO have comparable performances, whereas MGA has a higher median and standard deviation for the fitness function value for the multiple runs. The box and whisker plot for the folded cascode op-amp in Fig.~\ref{fig11}(b) shows that MABCO and MGWO have comparable performances and the median values obtained by MGA and MPSO are similar, but the standard deviation is higher for MGA. MABCO is found to perform consistently better with respect to the best value of the fitness function and the variation of the converged result across multiple runs, along with the mean run time for the two specific cases considered in this study. However, the results suggest that all the modified algorithms converge to a reasonable solution, with MABCO marginally better for the two cases considered. The results show that the modifications performed on the algorithms resulted in them exhibiting significant improvement in convergence and run-time for optimising the two op-amp topologies. The study shows that the proposed methodology can help design op-amp circuits with high accuracy and in less time. This study is performed with a fixed channel length for all the transistors. It can be easily extended to consider transistor lengths also as decision variables. The proposed approach with any of the presented modified algorithms can be extended for multi-objective optimisation problems and the design of more complex analog circuits.

 \section*{Acknowledgment}
 
 The authors acknowledge Ujjwal Rana for his contribution to programming the standard GWO algorithm and Krishna Kummarapalli for the fruitful discussions on analog IC design.

 \bibliographystyle{elsarticle-num} 
 \bibliography{main.bbl}

\begin{thebibliography}{10}
\expandafter\ifx\csname url\endcsname\relax
  \def\url#1{\texttt{#1}}\fi
\expandafter\ifx\csname urlprefix\endcsname\relax\def\urlprefix{URL }\fi
\expandafter\ifx\csname href\endcsname\relax
  \def\href#1#2{#2} \def\path#1{#1}\fi

\bibitem{1}
G.~Gielen, R.~Rutenbar, {Computer-aided design of analog and mixed-signal integrated circuits}, Proc. of the IEEE 88~(12) (2000) 1825--1854.
\newblock \href {https://doi.org/10.1109/5.899053} {\path{doi:10.1109/5.899053}}.

\bibitem{2}
C.~Toumazou, C.~Makris, {Analog IC design automation. I. Automated circuit generation: new concepts and methods}, IEEE Trans. Comput.-Aided Design Integr. Circuits Syst. 14~(2) (1995) 218--238.
\newblock \href {https://doi.org/10.1109/43.370422} {\path{doi:10.1109/43.370422}}.

\bibitem{3}
B.~Razavi, Design of Analog CMOS Integrated Circuits, 2nd Edition, McGraw-Hill Education, New York, USA, 2017.

\bibitem{4}
P.~Walker, J.~P. Ochoa-Ricoux, A.~Abusleme, Slice-based analog design, IEEE Access 9 (2021) 148164--148183.

\bibitem{5}
E.~Afacan, Inversion coefficient optimization based analog/rf circuit design automation, Microelectronics Journal 83 (2019) 86--93.

\bibitem{6}
S.~Yin, R.~Wang, J.~Zhang, Y.~Wang, Asynchronous parallel expected improvement matrix-based constrained multi-objective optimization for analog circuit sizing, IEEE Transactions on Circuits and Systems II: Express Briefs 69~(9) (2022) 3869--3873.

\bibitem{7}
M.~Hershenson, S.~Boyd, T.~Lee, {Optimal design of a CMOS op-amp via geometric programming}, IEEE Trans. Comput.-Aided Design Integr. Circuits Syst. 20~(1) (2001) 1--21.

\bibitem{8}
A.~Patanè, A.~Santoro, P.~Conca, G.~Carapezza, A.~L. Magna, V.~Romano, G.~Nicosia, Multi-objective optimization and analysis for the design space exploration of analog circuits and solar cells, Engineering Applications of Artificial Intelligence 62 (2017) 373--383.

\bibitem{9}
M.~N. Sabry, H.~Omran, M.~Dessouky, Systematic design and optimization of operational transconductance amplifier using gm/id design methodology, Microelectronics Journal 75 (2018) 87--96.

\bibitem{10}
M.~N. Sabry, I.~Nashaat, H.~Omran, Automated design and optimization flow for fully-differential switched capacitor amplifiers using recycling folded cascode ota, Microelectronics Journal 101 (2020) 104814.

\bibitem{11}
E.~Afacan, G.~Berkol, G.~Dundar, A.~E. Pusane, F.~Baskaya, An analog circuit synthesis tool based on efficient and reliable yield estimation, Microelectronics Journal 54 (2016) 14--22.

\bibitem{12}
A.~B. {de Andrade}, A.~Petraglia, C.~F. Soares, A constrained optimization approach for accurate and area efficient bandgap reference design, Microelectronics Journal 65 (2017) 72--77.

\bibitem{13}
J.~Olenšek, T.~Tuma, J.~Puhan, Árpád Bűrmen, A new asynchronous parallel global optimization method based on simulated annealing and differential evolution, Applied Soft Computing 11~(1) (2011) 1481--1489.

\bibitem{14}
G.~I. Tombak, Şeyda Nur~Güzelhan, E.~Afacan, G.~Dündar, {Simulated annealing assisted NSGA-III-based multi-objective analog IC sizing tool}, Integration 85 (2022) 48--56.

\bibitem{15}
Y.~Yang, H.~Zhu, Z.~Bi, C.~Yan, D.~Zhou, Y.~Su, X.~Zeng, {Smart-MSP: A Self-Adaptive Multiple Starting Point Optimization Approach for Analog Circuit Synthesis}, IEEE Trans. Comput.-Aided Design Integr. Circuits Syst. 37~(3) (2018) 531--544.

\bibitem{16}
C.~Vişan, O.~Pascu, M.~Stănescu, E.-D. Şandru, C.~Diaconu, A.~Buzo, G.~Pelz, H.~Cucu, Automated circuit sizing with multi-objective optimization based on differential evolution and bayesian inference, Knowledge-Based Systems 258 (2022) 109987.

\bibitem{17}
S.~Zhang, F.~Yang, C.~Yan, D.~Zhou, X.~Zeng, An efficient batch-constrained bayesian optimization approach for analog circuit synthesis via multiobjective acquisition ensemble, IEEE Transactions on Computer-Aided Design of Integrated Circuits and Systems 41~(1) (2022) 1--14.

\bibitem{18}
M.~Fayazi, Z.~Colter, E.~Afshari, R.~Dreslinski, {Applications of Artificial Intelligence on the Modeling and Optimization for Analog and Mixed-Signal Circuits: A Review}, IEEE Trans. Circuits Syst. I, Reg. Papers 68~(6) (2021) 2418--2431.

\bibitem{19}
A.~Budak, M.~Gandara, W.~Shi, D.~Pan, N.~Sun, B.~Liu, {An Efficient Analog Circuit Sizing Method Based on Machine Learning Assisted Global Optimization}, IEEE Trans. Comput.-Aided Design Integr. Circuits Syst. (2021) 1--1.

\bibitem{20}
B.~Liu, D.~Zhao, P.~Reynaert, G.~G.~E. Gielen, {Synthesis of Integrated Passive Components for High-Frequency RF ICs Based on Evolutionary Computation and Machine Learning Techniques}, IEEE Trans. Comput.-Aided Design Integr. Circuits Syst. 30~(10) (2011) 1458--1468.

\bibitem{21}
Y.~Li, Y.~Wang, Y.~Li, R.~Zhou, Z.~Lin, An artificial neural network assisted optimization system for analog design space exploration, IEEE Trans. Comput.-Aided Design Integr. Circuits Syst. 39~(10) (2020) 2640--2653.

\bibitem{22}
S.~Du, H.~Liu, Q.~Hong, C.~Wang, A surrogate-based parallel optimization of analog circuits using multi-acquisition functions, AEU - International Journal of Electronics and Communications 146 (2022) 154105.

\bibitem{23}
S.~Du, H.~Liu, H.~Yin, F.~Yu, J.~Li, A local surrogate-based parallel optimization for analog circuits, AEU - International Journal of Electronics and Communications 134 (2021) 153667.

\bibitem{24}
S.~Yin, R.~Wang, J.~Zhang, X.~Liu, Y.~Wang, Fast surrogate-assisted constrained multiobjective optimization for analog circuit sizing via self-adaptive incremental learning, IEEE Transactions on Computer-Aided Design of Integrated Circuits and Systems 42~(7) (2023) 2080--2093.

\bibitem{25}
M.~Dehbashian, M.~Maymandi-Nejad, An enhanced optimization kernel for analog ic design automation using the shrinking circles technique, Engineering Applications of Artificial Intelligence 58 (2017) 62--78.

\bibitem{26}
B.~Liu, Y.~Wang, Z.~Yu, L.~Liu, M.~Li, Z.~Wang, J.~Lu, F.~Fernandez, Analog circuit optimization system based on hybrid evolutionary algorithms, Integration, the VLSI Journal 42 (2009) 137--148.

\bibitem{27}
G.~Nicosia, S.~Rinaudo, E.~Sciacca, An evolutionary algorithm-based approach to robust analog circuit design using constrained multi-objective optimization, Knowledge-Based Systems 21~(3) (2008) 175--183, aI 2007.

\bibitem{28}
S.~Ghosh, B.~P. De, R.~Kar, D.~Mandal, A.~K. Mal, Optimal design of complementary metal-oxide-semiconductor analogue circuits: An evolutionary approach, Computers \& Electrical Engineering 80 (2019) 106485.

\bibitem{29}
R.~A. de~Lima~Moreto, C.~E. Thomaz, S.~P. Gimenez, A customized genetic algorithm with in-loop robustness analyses to boost the optimization process of analog cmos ics, Microelectronics Journal 92 (2019) 104595.

\bibitem{30}
A.~Jafari, E.~Bijami, H.~R. Bana, S.~Sadri, A design automation system for cmos analog integrated circuits using new hybrid shuffled frog leaping algorithm, Microelectronics Journal 43~(11) (2012) 908--915.

\bibitem{31}
R.~A. Vural, T.~Yildirim, T.~Kadioglu, A.~Basargan, Performance evaluation of evolutionary algorithms for optimal filter design, IEEE Trans. Evol. Comput. 16~(1) (2012) 135--147.

\bibitem{32}
M.~Fakhfakh, Y.~Cooren, A.~Sallem, M.~Loulou, P.~Siarry, {Analog circuit design optimization through the PSO technique}, Analog Integr. Circuits Signal Process. 63 (2010) 71--82.

\bibitem{33}
R.~A. Thakker, M.~S. Baghini, M.~B. Patil, Low-power low-voltage analog circuit design using hierarchical particle swarm optimization, in: 2009 22nd Int. Conf. on VLSI Design, 2009, pp. 427--432.

\bibitem{34}
R.~Vural, T.~Yildirim, Analog circuit sizing via swarm intelligence, AEU - International Journal of Electronics and Communications 66~(9) (2012) 732--740.

\bibitem{35}
B.~Benhala, Artificial bee colony technique for optimal design of folded cascode ota, in: 2018 Int. Conf. on Applied Mathematics Computer Science (ICAMCS), 2018, pp. 59--595.

\bibitem{36}
Y.~Delican, R.~Vural, T.~Yildirim, Artificial bee colony optimization based cmos inverter design considering propagation delays, in: 2010 XIth Int. Workshop on Symbolic and Numerical Methods, Modeling and Applications to Circuit Design (SM2ACD), 2010, pp. 1--5.

\bibitem{37}
D.~L. Wim~Kruiskamp, {DARWIN: CMOS opamp Synthesis by Means of a GA}, in: 32nd Design Automat. Conf., 1995, pp. 433--438.
\newblock \href {https://doi.org/10.1145/217474.217566} {\path{doi:10.1145/217474.217566}}.

\bibitem{38}
I.~Guerra-Gómez, E.~Tlelo-Cuautle, T.~McConaghy, G.~Gielen, Optimizing current conveyors by evolutionary algorithms including differential evolution, in: 2009 16th IEEE Int. Conf. on Electronics, Circuits and Systems - (ICECS 2009), 2009, pp. 259--262.

\bibitem{39}
M.~Taherzadeh-Sani, R.~Lotfi, H.~Zare-Hoseini, O.~Shoaei, Design optimization of analog integrated circuits using simulation-based genetic algorithm, in: Int. Symp. on Signals, Circuits and Systems, 2003. SCS 2003., Vol.~1, 2003, pp. 73--76 vol.1.

\bibitem{40}
H.~Xu, Y.~Ding, Optimizing method for analog circuit design using adaptive immune genetic algorithm, in: 2009 Fourth Int. Conf. on Frontier of Computer Science and Technology, 2009, pp. 359--363.

\bibitem{41}
R.~Zhou, P.~Poechmueller, Y.~Wang, An analog circuit design and optimization system with rule-guided genetic algorithm, IEEE Trans. Comput.-Aided Design Integr. Circuits Syst. 41~(12) (2022) 5182--5192.

\bibitem{42}
R.~{Rashid}, N.~Nambath, {Hybrid Particle Swarm Optimization Algorithm for Area Minimization in 65 nm Technology}, in: 2021 IEEE Int. Symp. on Circuits and Systems (ISCAS), 2021, pp. 1--5.

\bibitem{43}
R.~Rashid, N.~Nambath, {Area Optimisation of Two Stage Miller Compensated Op-Amp in 65 nm Using Hybrid PSO}, IEEE Trans. Circuits Syst., II, Exp. Briefs 69~(1) (2022) 199--203.

\bibitem{44}
B.~Karaboga, D.and~Basturk, {A powerful and efficient algorithm for numerical function optimization: artificial bee colony (ABC) algorithm}, J. Global Optimiz. 39~(3) (2007) 459–471.

\bibitem{45}
D.~Karaboga, B.~Akay, A comparative study of artificial bee colony algorithm, Applied Mathematics and Computation 214~(1) (2009) 108--132.

\bibitem{46}
J.~H. Holland, Adaptation in Natural and Artificial Systems: An Introductory Analysis with Applications to Biology, Control, and Artificial Intelligence, University of Michigan Press, Ann Arbor, 1975.

\bibitem{47}
S.~Mirjalili, Genetic Algorithm, Springer International Publishing, Cham, Switzerland, 2019, pp. 43--55.

\bibitem{48}
S.~Mirjalili, S.~M. Mirjalili, A.~Lewis, Grey wolf optimizer, Advances in Engineering Software 69 (2014) 46--61.

\bibitem{49}
J.~{Kennedy}, R.~{Eberhart}, {Particle Swarm Optimization}, in: Int. Conf. on Neural Networks, Vol.~4, 1995, pp. 1942--1948 vol.4.

\bibitem{50}
M.~Clerc, J.~Kennedy, The particle swarm - explosion, stability, and convergence in a multidimensional complex space, IEEE Trans. Evol. Comput. 6~(1) (2002) 58--73.

\bibitem{51}
Y.~Shi, R.~C. Eberhart, Parameter selection in particle swarm optimization, in: V.~W. Porto, N.~Saravanan, D.~Waagen, A.~E. Eiben (Eds.), Evolutionary Programming VII, Springer Berlin Heidelberg, Berlin, Heidelberg, 1998, pp. 591--600.

\bibitem{52}
A.~Lberni, M.~A. Marktani, A.~Ahaitouf, A.~Ahaitouf, Efficient butterfly inspired optimization algorithm for analog circuits design, Microelectronics Journal 113 (2021) 105078.

\bibitem{53}
P.~Allan~E., CMOS Analog Circuit Design, 3rd Edition, Oxford University Press, USA, 2012.

\bibitem{54}
B.~Liu, F.~V. Fern\'{a}ndez, G.~Gielen, R.~Castro-L\'{o}pez, E.~Roca, A memetic approach to the automatic design of high-performance analog integrated circuits, ACM Trans. Des. Autom. Electron. Syst. 14~(3) (jun 2009).

\bibitem{55}
A.~Sasikumar, V.~Subramaniyaswamy, R.~Jannali, V.~{Srinivasa Rao}, L.~Ravi, Design and area optimization of cmos operational amplifier circuit using hybrid flower pollination algorithm for iot end-node devices, Microprocessors and Microsystems 93 (2022) 104610.

\end{thebibliography}

\end{document}